\documentclass[journal]{IEEEtran}
\usepackage{graphicx}
\usepackage{color}
\usepackage{amssymb}
\usepackage{placeins}
\usepackage{float}
\usepackage{enumitem}
\usepackage{graphicx,epstopdf}
\usepackage{array}

\newenvironment{packed_enum}{
\begin{enumerate}[leftmargin=*]
  \setlength{\itemsep}{0.6pt}
  \setlength{\parskip}{0pt}
  \setlength{\parsep}{-2pt}
}{\end{enumerate}}

\usepackage{cite}

\usepackage[cmex10]{amsmath}

\usepackage{url}

\begin{document}
%
\title{Massive Online Crowdsourced Study of Subjective and Objective Picture Quality}

\author{Deepti~Ghadiyaram
        and~Alan~C. Bovik,~\IEEEmembership{Fellow,~IEEE}
\thanks{The authors are with the Laboratory for Image and Video Engineering, Department of Electrical and Computer Engineering, University of Texas at Austin, Austin, TX 78712 USA (e-mail: deepti@cs.utexas.edu; bovik@ece.utexas.edu).}}

\markboth{IEEE Transactions on Image Processing}%
{Ghadiyaram \MakeLowercase{\textit{et al.}}:Characterizing the Perception Quality of Real Distorted Images using Natural Scene Statistics and Deep Belief Nets.}
\maketitle

\begin{abstract}
Most publicly available image quality databases have been created under highly controlled conditions by introducing graded simulated distortions onto high-quality photographs. However, images captured using typical real-world mobile camera devices are usually afflicted by complex mixtures of multiple distortions, which are not necessarily well-modeled by the synthetic distortions found in existing databases. The originators of existing legacy databases usually conducted human psychometric studies to obtain statistically meaningful sets of human opinion scores on images in a stringently controlled visual environment, resulting in small data collections relative to other kinds of image analysis databases. Towards overcoming these limitations, we designed and created a new database that we call the \emph{LIVE In the Wild Image Quality Challenge Database}, which contains widely diverse authentic image distortions on a large number of images captured using a representative variety of modern mobile devices. We also designed and implemented a new online crowdsourcing system, which we have used to conduct a very large-scale, multi-month image quality assessment subjective study. Our database consists of over 350,000 opinion scores on 1,162 images evaluated by over 8100 unique human observers. Despite the lack of control over the experimental environments of the numerous study participants, we demonstrate excellent internal consistency of the subjective dataset. We also evaluate several top-performing blind IQA algorithms on it and present insights on how mixtures of distortions challenge both end users as well as automatic perceptual quality prediction models. The new database is available for public use at \url{https://www.cs.utexas.edu/~deepti/ChallengeDB.zip}.
\end{abstract}


%
\IEEEpeerreviewmaketitle

\section{Introduction}
\IEEEPARstart{T}he field of visual media has been witnessing explosive growth in recent years, driven by significant advances in technology that have been made by camera and mobile device manufacturers, and by the synergistic development of very large photo-centric social networking websites, which allow consumers to efficiently capture, store, and share high-resolution images with their friends or the community at large. The vast majority of these digital pictures are taken by casual, inexpert users, where the capture process is affected by delicate variables such as lighting, exposure, aperture, noise sensitivity, and lens limitations, each of which could perturb an image's perceived visual quality. Though cameras typically allow users to control the parameters of image acquisition to a certain extent, the unsure eyes and hands of most amateur photographers frequently lead to occurrences of annoying image artifacts during capture. This leads to large numbers of images of unsatisfactory perceptual quality being captured and stored along with more desirable ones. Being able to automatically identify and cull low quality images, or to prevent their occurrence by suitable quality correction processes during capture are thus highly desirable goals that could be enabled by automatic quality prediction tools \cite{bovik13}. Thus, the development of objective image quality models from which accurate predictions of the quality of digital pictures as perceived by human observers can be derived has greatly accelerated. 

Advances in practical methods that can efficiently predict the perceptual quality of images have the potential to significantly impact protocols for monitoring and controlling multimedia services on wired and wireless networks and devices. These methods have the potential to also improve the quality of visual signals by acquiring or transporting them via ``quality-aware'' processes. Such ``quality-aware'' processes could perceptually optimize the capture process and modify transmission rates to ensure good quality across wired or wireless networks. Such strategies could help ensure that end users have a satisfactory quality of experience (QoE).

The goal of an objective no-reference image quality assessment (NR IQA) model is as follows: given an image (possibly distorted) and no other additional information, automatically and accurately predict its perceptual quality. Given that the ultimate receivers of these images are humans, the only reliable way to understand and predict the effect of distortions on a typical person's viewing experience is to capture opinions from a large sample of human subjects, which is termed \textit{subjective image quality assessment}. While these subjective scores are vital for understanding human perception of image quality, they are also crucial for designing and evaluating reliable IQA models that are consistent with subjective human evaluations, regardless of the type and severity of the distortions. 

The most efficient NR IQA algorithms to date are founded on the statistical properties of natural\footnote{Natural images are not necessarily images of natural environments such as trees or skies. Any natural visible-light image that is captured by an optical camera and is not subjected to artificial processing on a computer is regarded here as a natural image including photographs of man-made objects.} images. Natural scene statistics (NSS) models \cite{bovik13} are based on the well-established observation that good quality real-world photographic images\footnote{We henceforth refer to such images as `pristine' images.} obey certain \emph{perceptually} relevant statistical laws that are violated by the presence of common image distortions. Some state-of-the-art NR IQA models \cite{brisque, diviine, bliinds2, lbiq, cbiq, niqe, tang-cvpr, cdiivine, s3Index} that are based on NSS models attempt to quantify the degree of `naturalness' or `unnaturalness' of images by exploiting these statistical perturbations. This is also true of competitive reduced-reference IQA models \cite{rred}. Such statistical `naturalness' metrics serve as image features which are typically deployed in a supervised learning paradigm, where a kernel function is learned to map the features to ground truth subjective quality scores. A good summary of such models and their quality prediction performance can be found in \cite{brisque}. 

\begin{figure*}[t]
\begin{center}$
\begin{array}{cccc}
\includegraphics[width=1.3in,height=2.0in]{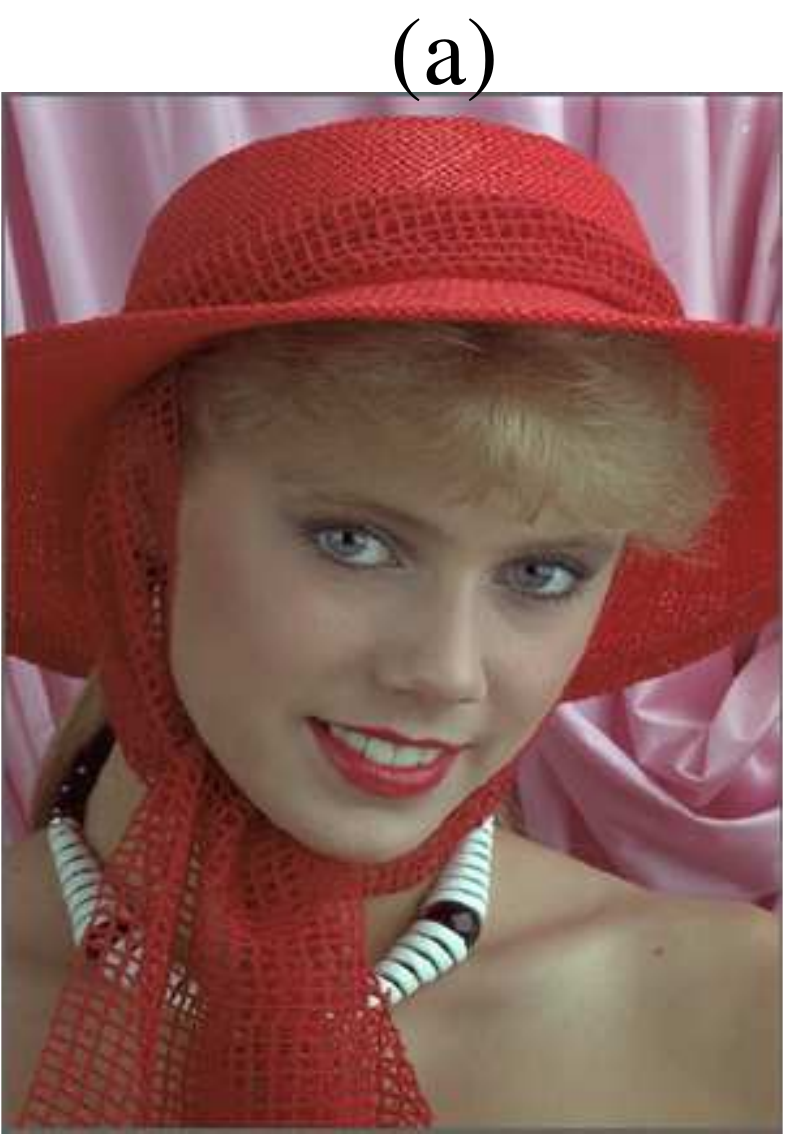} &
\includegraphics[width=1.3in,height=2.0in]{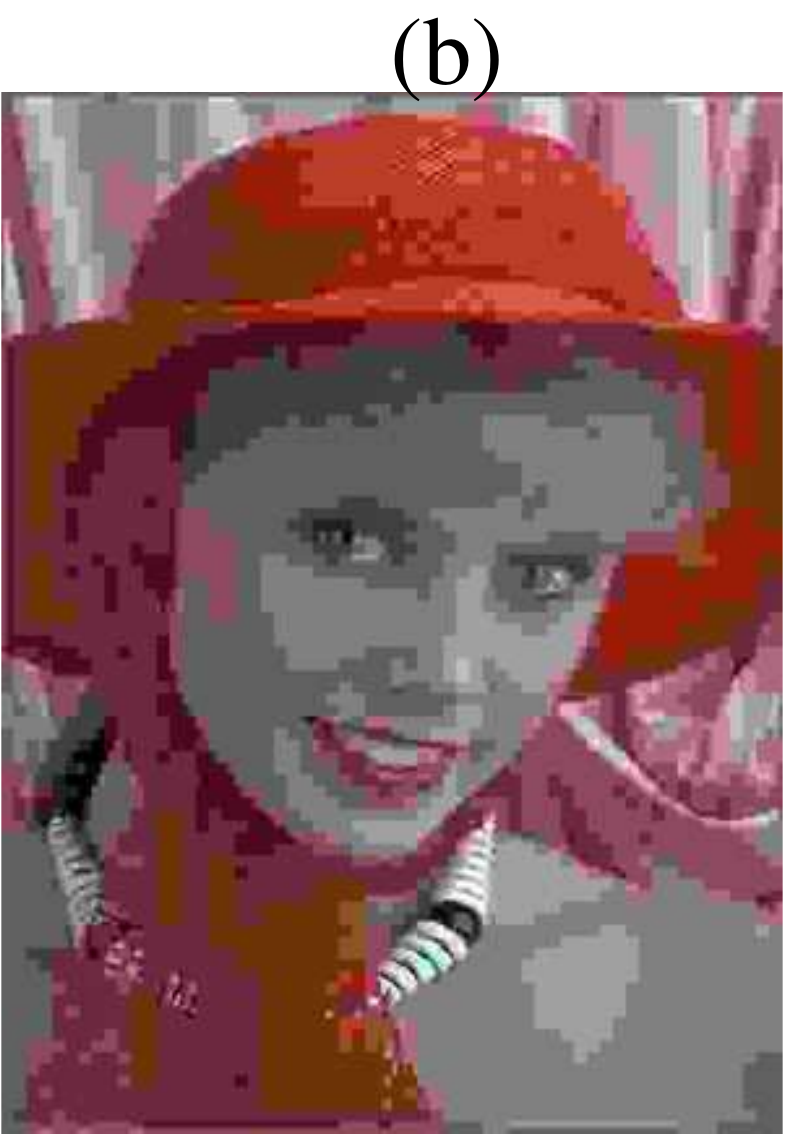} &
\includegraphics[width=1.3in,height=2.0in]{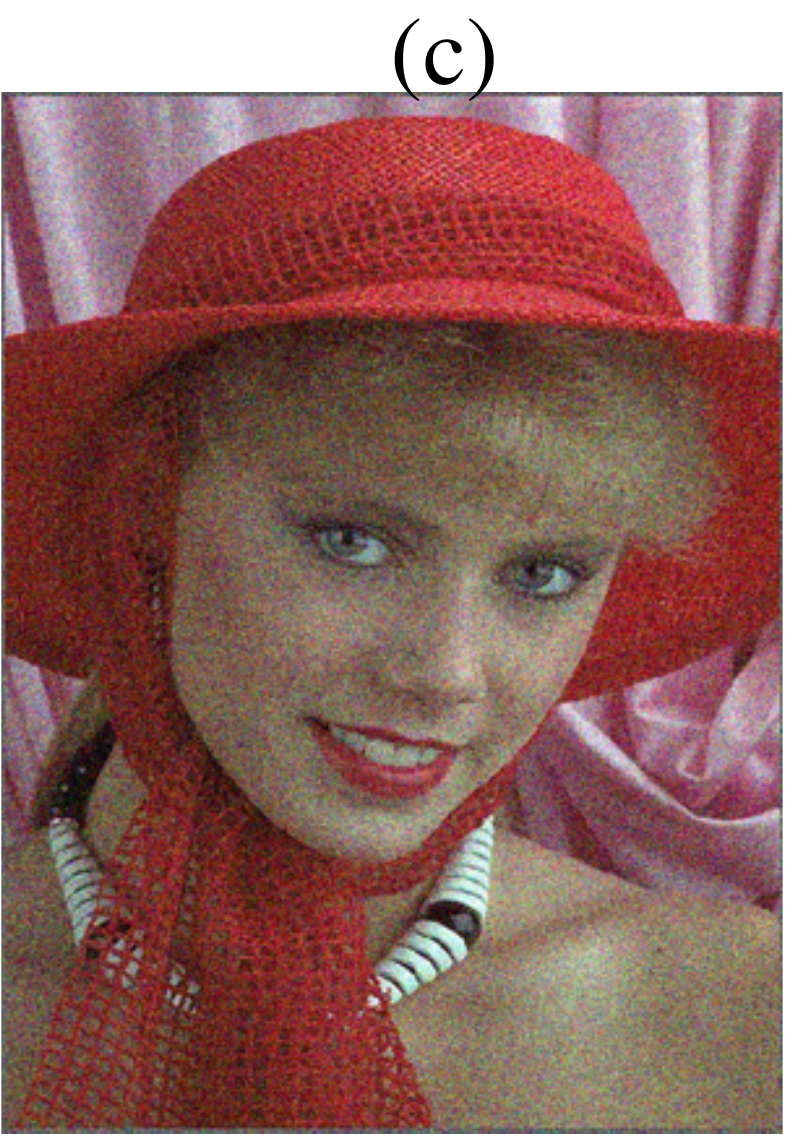} &
\includegraphics[width=1.5in,height=2.0in]{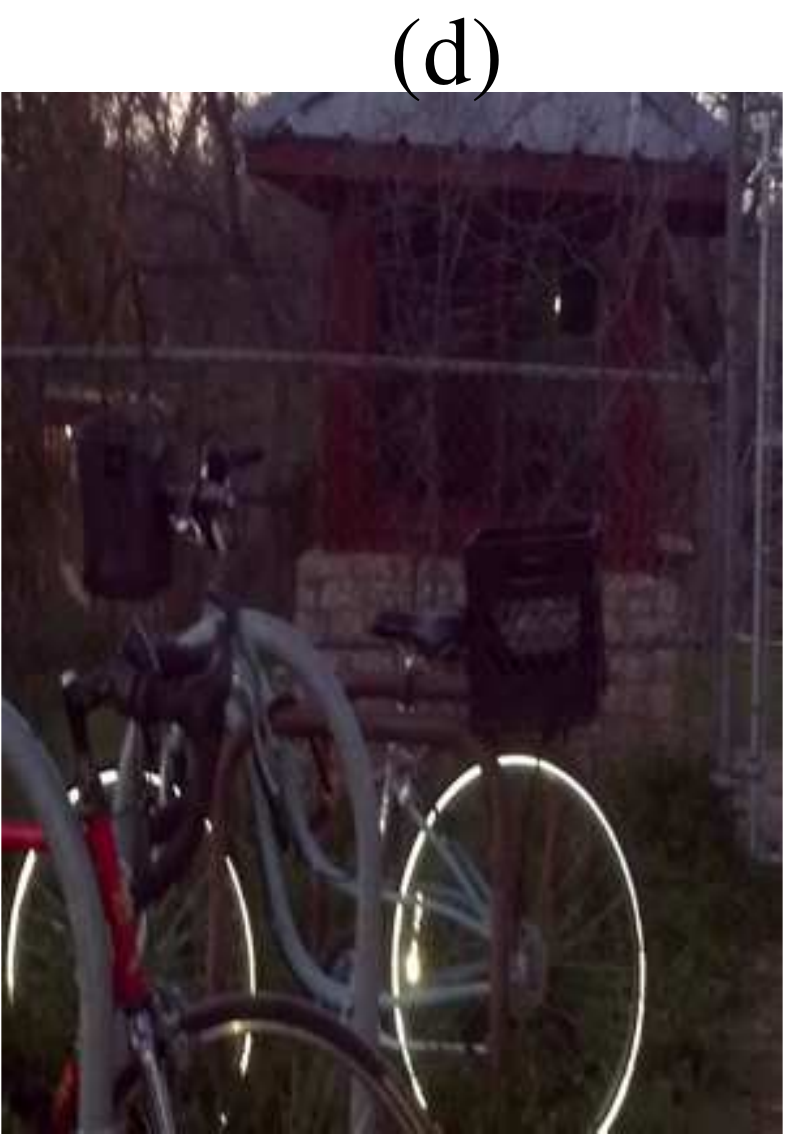}
\end{array}$ 
\caption{\small{(a) A pristine image from the legacy LIVE Image Quality Database \cite{live-r2} (b) JPEG compression distortion artificially applied to (a). (c) White noise added to (a). (d) A blurry image also distorted with low-light noise from the new LIVE In the Wild Image Quality Challenge Database. }}
\label{sampleImgs}
\end{center}
\end{figure*}

\textbf{Authentic distortions:} Current blind IQA models \cite{brisque, diviine, bliinds2, lbiq, cbiq, niqe, tang-cvpr, cdiivine, s3Index} use legacy benchmark databases such as the LIVE Image Quality Database \cite{live-r2} and the TID2008 Database \cite{tid} to train low-level statistical image quality cues against recorded subjective quality judgements. These databases, however, have been designed to contain images corrupted by only one of a few synthetically introduced distortions, e.g., images containing only JPEG compression artifacts, images corrupted by simulated camera sensor noise, or by simulated blur. Though the existing legacy image quality databases have played an important role in advancing the field of image quality prediction, we contend that determining image quality databases such that the distorted images are \emph{derived} from a set of high-quality source images and by simulating image impairments on them is much too limiting. In particular, traditional databases fail to account for difficult mixtures of distortions that are inherently introduced during image acquisition and subsequent processing and transmission. For instance, consider the images shown in Fig. \ref{sampleImgs}(a) - Fig. \ref{sampleImgs}(d). Figure \ref{sampleImgs}(d) was captured using a mobile device and can be observed to be distorted by both low-light noise and compression errors. Figure 1(b) and (c) are from the legacy LIVE IQA Database \cite{live-r2} where JPEG compression and Gaussian blur distortions were synthetically introduced on a pristine image (Fig. \ref{sampleImgs}(a)).

Although such singly distorted images (and datasets) facilitate the study of the effects of distortion-specific parameters on human perception, they omit important and frequently occurring mixtures of distortions that occur in images captured using mobile devices. This limitation is especially problematic for \emph{blind} IQA models which have great potential to be employed in large-scale user-centric visual media applications. Designing, training, and evaluating IQA models based only on the statistical perturbations observed on these restrictive and non-representative datasets might result in quality prediction models that inadvertently assume that every image has a ``single'' distortion that most objective viewers could agree upon. Although top-performing algorithms perform exceedingly well on these legacy databases (e.g., the median Spearmann correlation of $0.94$ on the legacy LIVE IQA Database \cite{live-r2} reported by BRISQUE \cite{brisque} and $0.96$ reported by Tang \emph{et. al} in \cite{tang-cvpr}), their performance is questionable when tested on naturally distorted images that are normally captured using mobile devices under highly variable illumination conditions. Indeed, we will show in Sec. \ref{sec:expt} that the performance of several top-performing algorithms staggers when tested on images corrupted by diverse authentic and mixed, multipartite distortions such as those contained in the new LIVE In the Wild Image Quality Challenge Database.

With this in mind, we formulated the goal of designing a unique and challenging database containing a large number of \emph{authentically} distorted images. Of course, we had to consider the question: What makes an image quality database representative? Beyond capturing a wide variety of image content including pictures of people, objects, and both indoor and outdoor scenes, such a database should also account for the various and diverse frequently used commercial image capture devices. The corpus of pictures should also have been obtained under varied illumination conditions and, given the propensity of users to acquire their pictures in imperfect ways, they should exhibit a broad spectrum of authentic quality ``types,'' mixtures, and distortion severities. The result of our effort is a difficult new resource called the \textbf{LIVE In the Wild Image Quality Challenge Database}. Of course, since we did not synthetically distort images, no pristine references are available; hence the new database is only suitable for no-reference IQA research.

\textbf{Large-scale subjective study:} Since existing records of human quality judgements are associated with legacy ``synthetic'' databases, another important contribution that we made is to acquire human opinion scores on authentically distorted images. Most existing no-reference image quality assessment models follow a supervised learning paradigm, hence their accuracy is strongly linked to the quality and quantity of the annotated training data available. Having access to more ratings per image means that variations of perceived quality can be more easily captured. The human opinion scores in most of the legacy datasets were collected by conducting subjective studies in fixed laboratory setups, where images were displayed on a single device with a fixed display resolution and which the subjects viewed from a fixed distance. However, significant advances in technology made by camera and mobile device manufacturers now allow users to efficiently access visual media over wired and wireless networks. Thus, the subjective image quality opinions gathered under artificially controlled settings do not necessarily mirror the picture quality perceived on widely used portable display devices having varied resolutions. Gathering representative subjective opinions by simulating different viewing conditions would be exceedingly time-consuming, cumbersome, and would require substantial manual effort. On the other hand, exploring novel ways to collect subjective scores online requires dealing with noisy data, establishing the reliability of the obtained human opinions, etc. - a challenging research topic. 

As we describe in the following, we have conducted such a study to obtain a very large number of human opinion scores on the LIVE Challenge image data using a sophisticated crowdsourcing system design. We explain the human study, the method of validating the data obtained, and demonstrate how it can be used to fruitfully advance the field of no-reference/blind image quality prediction. 



\textbf{Contributions:} We describe our attack on the difficult problem of blind image quality assessment on authentically distorted images from the ground up and summarize our contributions below:
\begin{packed_enum}
\item{First, we introduce the content and characteristics of the new LIVE In the Wild Image Quality Challenge Database, which contains $1162$ authentically distorted images captured from many diverse mobile devices. Each image was collected without artificially introducing any distortions beyond those occurring during capture, processing, and storage by a user's device.}
\item{Next, we aimed to gather very rich human data, so we designed and implemented an extensive online subjective study by leveraging Amazon's crowdsourcing system, the Mechanical Turk. We will describe the design and infrastructure of our online crowdsourcing system\footnote{A report describing early progress of this work appeared in \cite{studyPaper}.} and how we used it to conduct a very large-scale, multi-month image quality assessment subjective study, wherein a wide range of diverse observers recorded their judgments of image quality.}
\item{We also discuss the critical factors that are involved in successfully crowdsourcing human IQA judgments, such as the overall system design of the online study, methods for subject validation and rejection, task remuneration, influence of the subjective study conditions on end users' assessment of perceptual quality, and so on.}
\item{As a demonstration of the usefulness of the study outcomes, we also conducted extensive empirical studies on the performance of several top-performing NR IQA models (Sec. \ref{sec:expt}), both on a legacy benchmark dataset \cite{live-r2} as well as on the new LIVE In the Wild Image Quality Challenge Database.}
\end{packed_enum}
 Our results demonstrate that: (1) State-of-the-art NR IQA algorithms perform poorly on the LIVE In the Wild Image Quality Challenge Database, which has a high percentage of images distorted by multiple processes, all of which are authentic (Sec. \ref{sec:diffTech}). (2) Our human-powered, crowdsourcing framework proved to be an effective way to gather a large number of opinions from a diverse, distributed populace over the web. So far, we have collected over $350,000$ human opinion scores on 1,162 naturally distorted images from over $8,100$ distinct subjects, making it the world's largest, most comprehensive online study of perceptual image quality ever conducted. (3) A correlation of $0.9851$ was obtained between the MOS values gathered from the proposed crowdsourcing platform and those from a different study conducted by the creators of the LIVE Multiply Distorted Image Quality Database \cite{multiply}. This high correlation advocates the veracity of our designed online system in gathering reliable human opinion scores (Sec. \ref{goldStandard}). 

\section{Related Work}
\textbf{Benchmark IQA Databases - Content and Test Methodologies:} \label{sec:databases}
Most of the top-performing IQA models (full, reduced, and no-reference) have been extensively evaluated on two benchmark databases: the LIVE IQA Database which was designed in $2005$ and the TID2008 Database, designed and released in $2008$. The LIVE IQA Database, one of the first comprehensive IQA databases, consists of $779$ images, much larger than the small databases that existed at the time of its introduction \cite{sheikh-1} -\cite{sheikh-3}. This legacy database contains $29$ pristine reference images and models five distortion types - jp2k, jpeg, Gaussian blur, white noise, and fast fading noise \cite{live-r2}. The TID2008 Database is larger, consisting of $25$ reference and $1700$ distorted images over $17$ distortion categories. TID2013 \cite{tid2013} is a very recently introduced image quality database with an end goal to include the peculiarities of color distortions in addition to the 17 simulated spatial distortions included in TID2008. It consists of 3000 images and includes seven new types of distortions, thus modeling a total of 24 distortions. We refer the reader to \cite{tid, tid2013} for more details on the categories and severities of image distortions contained in this database.
\begin{figure*}[t] 
\begin{center}
\includegraphics[width=8cm]{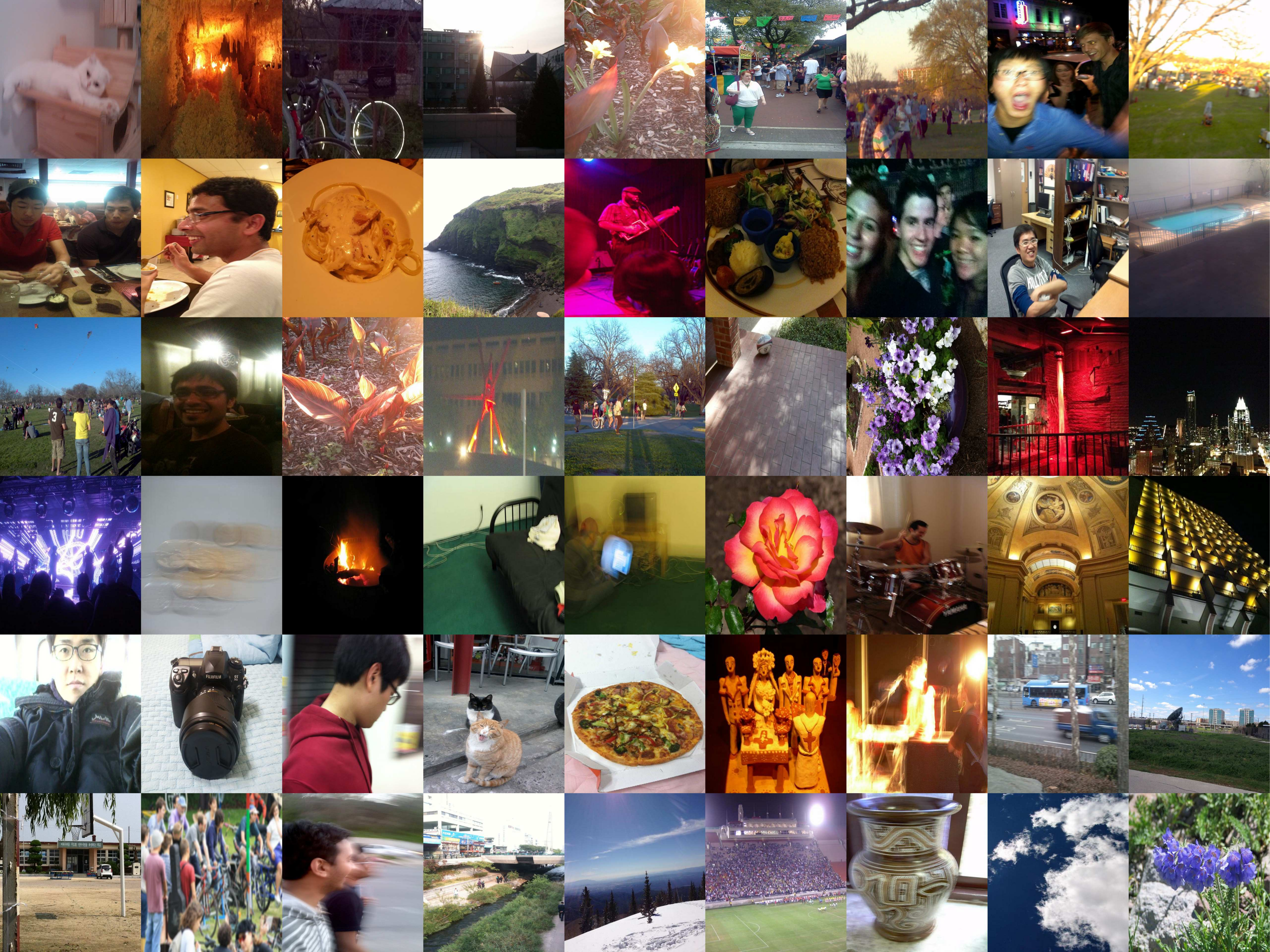}
\caption{\small{Sample images from the LIVE In the Wild Image Quality Challenge Database. These images include pictures of faces, people, animals, close-up shots, wide-angle shots, nature scenes, man-made objects, images with distinct foreground/background configurations, and images without any specific object of interest.}}
\label{fig:challengeImgs}
\end{center}
\end{figure*}

These databases \cite{live-r2, tid, tid2013} contain quality ratings obtained by conducting subjective studies in controlled laboratory settings\footnote{The authors of \cite{tid} report that about 200 of their observers have participated in the study via the Internet.}. The TID2008 opinion scores were obtained from $838$ observers by conducting batches of large scale subjective studies, whereby a total of $256,000$ comparisons of the visual quality of distorted images were performed. Although this is a large database, some of the test methodologies that were adopted do not abide by the ITU recommendations. For instance, the authors followed a \emph{swiss competition principle} and presented three images, wherein two of them are the distorted versions of the third one. A subject was asked to choose one image of superior quality amongst the two distorted images. We believe that this kind of presentation does not accurately reflect the experience of viewing and assessing distorted images in the most common (e.g. mobile) viewing scenarios. Furthermore, in each experiment, a subject would view and compare $306$ instances of the same reference image containing multiple types and degrees of distortions, introducing the significant possibility of serious \emph{hysteresis effects} that are not accounted for when processing the individual opinion scores. 

In pairwise comparison studies, the method for calculating preferential ranking of the data can often dictate the reliability of the results. Certain probabilistic choice model-based ranking approaches \cite{ranking1, ranking2, ranking3} offer sophisticated ways to accurately generate quality rankings of images. However, the opinion scores in the TID2008 database were obtained by first accumulating the points ``won'' by each image. These points are driven by the preferential choices of different observers during the comparative study. The mean values of the winning points on each image were computed in the range $[0-9]$ and are referred to as mean opinion scores. This method of gathering opinion scores, which diverges from accepted practice, is in our view questionable. 

Conversely, the LIVE IQA Database was created following an ITU recommended single-stimulus methodology. Both the reference images as well as their distorted versions were evaluated by each subject during each session. Thus, \emph{quality difference scores} which address user biases were derived for all the distorted images and for all the subjects. Although the LIVE test methodology and subject rejection method adheres to the ITU recommendations, the test sessions were designed to present a subject with a set of images, all afflicted by the same type of distortion (for instance, all the images in a given session consisted of different degrees of JPEG 2000 distortion) that were artificially added to different reference images. We suspect that this could have led to over-learning of each distortion type by the subjects as the study session progressed.

Since cameras on mobile devices make it extremely easy to snap images spontaneously under varied conditions, the complex mixtures of image distortions that occur are not well-represented by the distorted image content in either of these legacy image databases. This greatly motivated us to acquire real images suffering the natural gamut of authentic distortion mixtures as the basis for a large new database and human study. Such a resource could prove quite valuable for the design of next-generation robust IQA prediction models that will be used to ensure future end users' high quality of viewing experience.

\textbf{Online Subjective Studies} \label{sec:onlineStudies}
Most subjective image quality studies have been conducted in laboratory settings with stringent controls on the experimental environment and involving small, non-representative subject samples (typically graduate and undergraduate university students). For instance, the creators of the LIVE IQA Database used two 21-inch CRT monitors with display resolutions of $1024 \times 768$ pixels in a normally lit room, which the subjects viewed from a viewing distance of $2 - 2.5$ screen heights.

However, the highly variable ambient conditions and the wide array of display devices on which a user might potentially view images will have a considerable influence on her perception of picture quality. This greatly motivates our interest in conducting IQA studies on the Internet, which can enable us to access a much larger and more diverse subject pool while allowing for more flexible study conditions. However, the lack of control on the subjective study environment introduces several challenges (more in Sec. \ref{crowdsourceChallenges}), some of which can be handled by employing counter measures (such as gathering details of the subject's display monitor, room illumination, and so on) \cite{hossfeld-stall}.

A few studies have recently been reported that used web-based image, video, or audio rating platforms \cite{QoE , siq, pixelJoust, colorwars, crowdMOS, grzy, qualityCrowd, inmomento}. Some of these studies employed pairwise comparisons followed by ranking techniques \cite{ranking1, ranking2, ranking3} to derive quality scores, while others adopted the single stimulus technique and an absolute category rating (ACR) scale. Since performing a complete set of paired comparisons (and ranking) is time-consuming and monetarily expensive when applied on a large scale, Xu \textit{et al.} \cite{xu1, xu2} introduced the HodgeRank on Random Graphs (HRRG) test, where random sampling methods based on Erd{\"o}s-R{\'e}nyi random graphs were used to sample pairs and the HodgeRank \cite{hodgerank} was used to recover the underlying quality scores from the incomplete and imbalanced set of paired comparisons. More recently, an active sampling method \cite{doerman} was proposed that actively constructs a set of queries consisting of single and pair-wise tests based on the expected information gain provided by each test with a goal to reduce the number of tests required to achieve a target accuracy. However, all of these studies were conducted on small sets of images taken from publicly available databases of synthetically distorted images \cite{live-r2}, mostly to study the reliability and quality of the opinion scores obtained via the Internet testing methodology. In most cases, the subjective data from these online studies is publicly unavailable.

To the best of our knowledge, we are aware of only one other project \cite{crowdMOS} reporting efforts made in the same spirit as our work, that is, \emph{crowdsourcing} the image subjective study on Mechanical Turk by following a single-stimulus methodology\footnote{The authors of \cite{QoE} also used Mechanical Turk, but they adopted a pairwise comparison methodology.}. However, the authors of \cite{crowdMOS} tested their crowdsourcing system on only $116$ JPEG compressed images from the legacy LIVE Image Quality Database of synthetically distorted images \cite{live-r2} and gathered opinion scores from only forty subjects. By contrast, the new LIVE In the Wild Image Quality Challenge Database has $1162$ challenging images and engaged more than $8100$ unique subjects. Also, we wanted our web-based online study to be similar to the subjective studies conducted under laboratory settings with instructions, training, and test phases (more details in Sec. \ref{sec:inst}). We also wanted unique participants to view and rate the images on a continuous rating scale (as opposed to using the ACR scale). Thus we chose to design our own crowdsourcing framework incorporating all of the above design choices, as none of the existing successful crowdsourcing frameworks \cite{QoE, crowdMOS, qualityCrowd, inmomento, wesp} seemed to offer us the flexibility and control that we desired.

\section{LIVE In the Wild Image Quality Challenge Database}
 In practice, every image captured by a typical mobile digital camera device passes several processing stages, each of which can introduce visual artifacts. Authentically distorted images captured using modern cameras are likely to be impaired by sundry and mixed artifacts such as low-light noise and blur, motion-induced blur, over and underexposure, compression errors, and so on.

\begin{figure}[t]
\begin{center}
\includegraphics[height=1.1in]{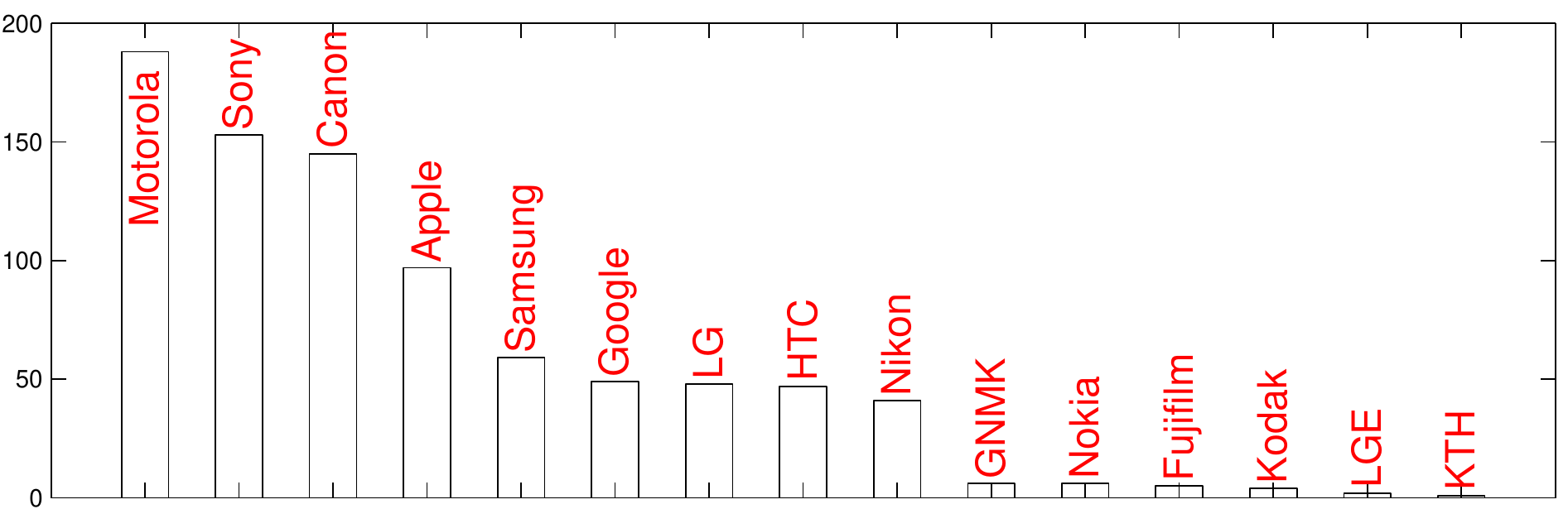}
\caption{\small{Distribution of different manufacturers of the cameras that were used to capture a sample of images contained in our database.}}
\label{fig:deviceDist}
\end{center}
\end{figure} 

The lack of content diversity and mixtures of bonafide distortions in existing, widely-used image quality databases \cite{tid, live-r2} is a continuing barrier to the development of better IQA models and prediction algorithms of the perception of real-world image distortions. To overcome these limitations and towards creating a holistic resource for designing the next generation of robust, perceptually-aware image assessment models, we designed and created the LIVE In the Wild Image Quality Challenge Database, containing images afflicted by diverse authentic distortion mixtures on a variety of commercial devices. 

Figure \ref{fig:challengeImgs} presents a few images from this database. The images in the database were captured using a wide variety of mobile device cameras as shown in Fig. \ref{fig:deviceDist}. The images include pictures of faces, people, animals, close-up shots, wide-angle shots, nature scenes, man-made objects, images with distinct foreground/background configurations, and images without any specific object of interest. Some images contain high luminance and/or color activity, while some are mostly smooth. Since these images are naturally distorted as opposed to being artificially distorted post-acquisition pristine reference images, they often contain mixtures of multiple distortions creating an even broader spectrum of perceivable impairments.

\subsection{Distortion Categories}\label{sec:distCategories}
\begin{figure*}[t]
\begin{center}
\includegraphics[width=5.5in, height=3.3in]{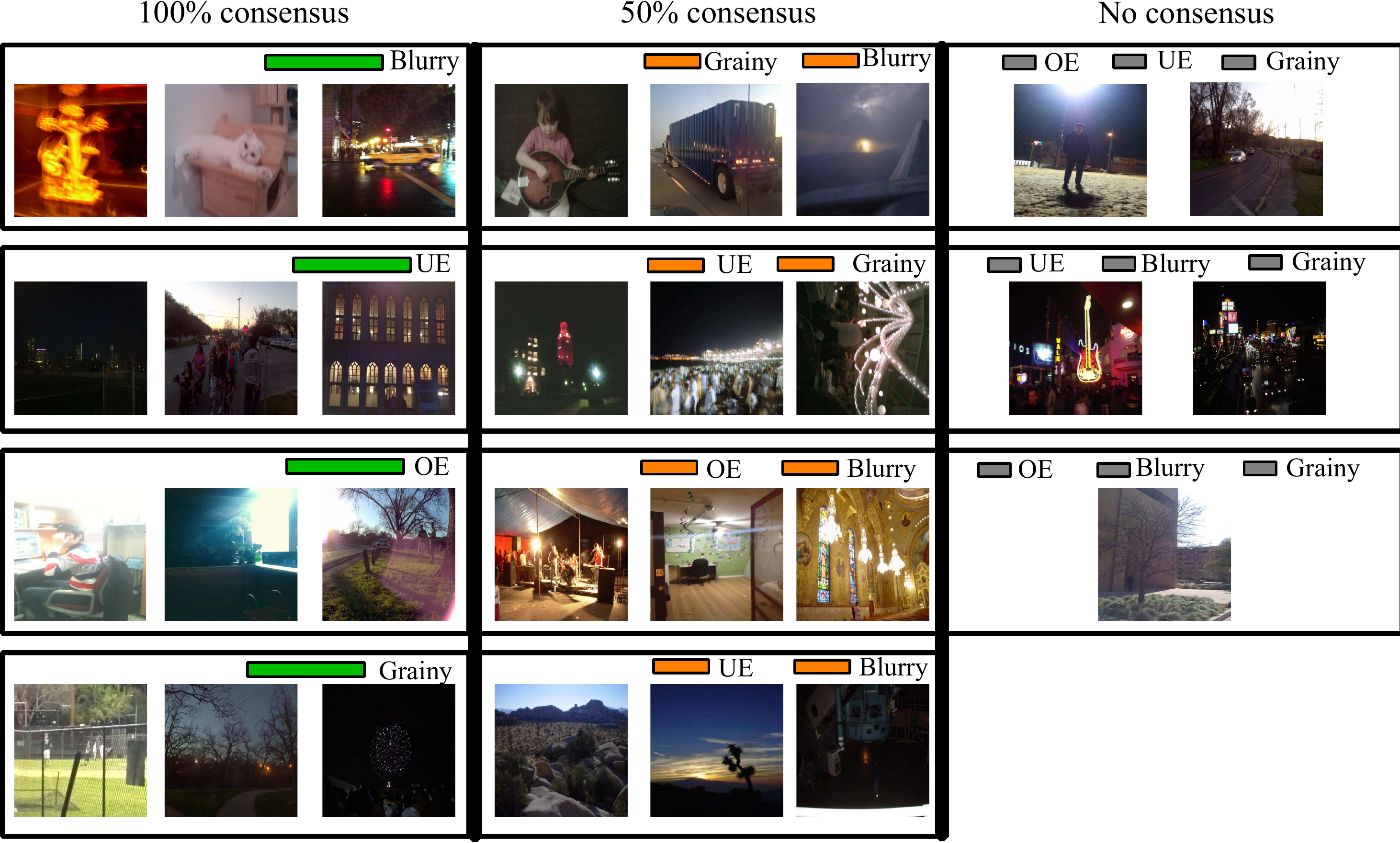}
\caption{\small{Different distortion category labels obtained after aggregating the data. The images presented in the left column mostly had one dominant distortion that most subjects could agree opon. The images presented in the next two columns have multiple distortions leading to disagreement amongst the opinions of the subjects (OE, UE = Over-, Underexposed.) Best viewed in color.}}
\label{fig:categoryClashes}
\end{center}
\end{figure*}

Since the images in our database contain mixtures of unknown distortions, in addition to gathering perceptual quality opinion scores on them (as discussed in detail in Sec. \ref{sec:onlinecrowdsource}), we also wanted to understand to what extent the subjects could supply a sense of distortion type against a few categories of common impairments. Thus we also conducted a \underline{\textbf{separate}} crowdsourcing study wherein the subjects were asked to select the single option from among a list of distortion categories that they think represented the \emph{most dominant distortion in each presented image}. The categories available to choose from were - ``Blurry,'' ``Grainy,'' ``Overexposed,'' ``Underexposed,'' ``No apparent distortion. Its a great image,'' and ``I don't understand the question.'' We adopted a majority voting policy to aggregate the distortion category labels obtained on every image from several subjects. A few images along with the category labels gathered on them are shown in Fig. \ref{fig:categoryClashes}. 

\subsubsection{100\% consensus:} Images presented in the left column of Fig. \ref{fig:categoryClashes} were sampled from an image pool where a majority of the subjects were in full agreement with regard to their opinion of the specific distortion present in those images. 

\subsubsection{50\% consensus:} The images presented in the second column are from a pool of images that received an approximately equal number of votes for two different classes of distortions. That is, about 50\% of the subjects who viewed these images perceived one kind of dominant distortion while the remaining subjects perceived a completely different distortion to be the most dominating one. 

\subsubsection{No consensus:} The confusion of choosing a dominant distortion was more difficult for some images, a few of which are presented in the last column. Here, nearly a third of the total subjects who were presented with these images labeled them as belonging to a distortion category different from the two other dominant labels obtained from the other subjects. 

Figure \ref{fig:categoryClashes} highlights the risk of forcing a consensus on image distortion categories through majority voting on our dataset. Multiple objective viewers appeared to have different sensitivities to different types of distortions which, in combination with several other factors such as display device, viewer distance from the screen, and image content, invariably affect his/her interpretation of the underlying image distortion. This non-negligible disagreement among human annotators sheds light on the extent of distortion variability and the difficulty of the data contained in the current database. We hope to build on these insights to develop a holistic identifier of mixtures of authentic distortions in the near future. For now, we take this as more direct evidence of the overall complexity of the problem. 
\\
\\
\textbf{No well-defined distortion categories in real-world pictures:} The above study highlights an important characteristic of real-world, authentically distorted images captured by na{\"i}ve users of consumer camera devices - that these pictures cannot be accurately described as generally suffering from \emph{single} distortions. Normally, inexpert camera users will acquire pictures under highly varied illumination conditions, with unsteady hands, and with unpredictable behavior on the part of the photographic subjects. Further, the overall distortion of an image also depends on other factors such as device and lens configurations. Furthermore, authentic mixtures of distortions are even more difficult to model when they interact, creating new agglomerated distortions not resembling any of the constituent distortions. Indeed real-world images sufer from a many-dimensional continuum of distortion perturbations. For this reason, it is not meaningful to attempt to segregate the images in the LIVE In the Wild Image Quality Challenge Database into discrete distortion categories.

\section{Crowdsourced framework for gathering subjective scores}\label{sec:onlinecrowdsource}
Crowdsourcing systems like Amazon Mechanical Turk (AMT), Crowd Flower \cite{crowdflower}, and so on, have emerged as effective, human-powered platforms that make it feasible to gather a large number of opinions from a diverse, distributed populace over the web. On these platforms, ``requesters'' broadcast their task to a selected pool of registered ``workers'' in the form of an open call for data collection. Workers who select the task are motivated primarily by the monetary compensation offered by the requesters and also by the enjoyment they experience through participation.

\subsection{Challenges of Crowdsourcing} \label{crowdsourceChallenges}
Despite the advantages offered by crowdsourcing frameworks, there are a number of well-studied limitations of the same. For example, requesters have limited control over the study setup and on factors such as the illumination of the room and the display devices being used by the workers. Since these factors could be relevant to the subjective evaluation of perceived image quality, we gathered information on these factors in a compulsory survey session presented towards the end of the study (more details in Sec. \ref{sec:inst}).

The basic study structure and procedures of subjective testing in a crowdsourcing framework differ from those of traditional subjective studies conducted in a laboratory. Subjective tests conducted in a lab environment typically last for many minutes with a goal of gathering ratings on every image in the dataset and are usually conducted in multiple sessions to avoid subject fatigue. For instance, the study reported in \cite{live-r2} was conducted in two sessions where each session lasted for 30 minutes. However, crowdsourced tasks should be small enough that they can be completed by workers quickly and with ease. It has been observed that it is difficult to find workers to participate in large and more time consuming tasks, since many workers prefer high rewards per hour \cite{task-properties}. Thus, an online test needs to be partitioned into smaller chunks. Further, although requesters can control the maximum number of tasks each worker can participate in, they cannot control the exact number of times a worker selects a task. Thus, it is very likely that all the images in the dataset will not be viewed and rated by every participating worker.

Despite these limitations imposed on any crowdsourcing framework, our online subjective study, which we describe in great detail below has enabled us to gather a large number of highly reliable opinion scores on all the images in our dataset.

Image aesthetics are closely tied to perceived quality and crowdsourcing platforms have been used in the past to study the aesthetic appeal of images \cite{aesthetics}. Here we have focused on gathering subjective quality scores using highly diverse aesthetic content. We also informed users how to focus on quality and not aesthetics. In future studies, it will be of value to gather associated side information from each subject regarding the content and aesthetics of each presented image (or video).

\subsection{Instructions, Training, and Testing} \label{sec:inst}
The data collection tasks on AMT are packaged as HITs (Human Intelligence Tasks) by requesters and are presented to workers, who first visit an instructions page which explains the details of the task. If the worker understands and likes the task, she needs to click the ``Accept HIT'' button which then directs her to the actual task page at the end of which, she clicks a ``Submit Results'' button for the requester to capture the data. 

\begin{figure*}[t]
\begin{center}
\includegraphics[width=9cm]{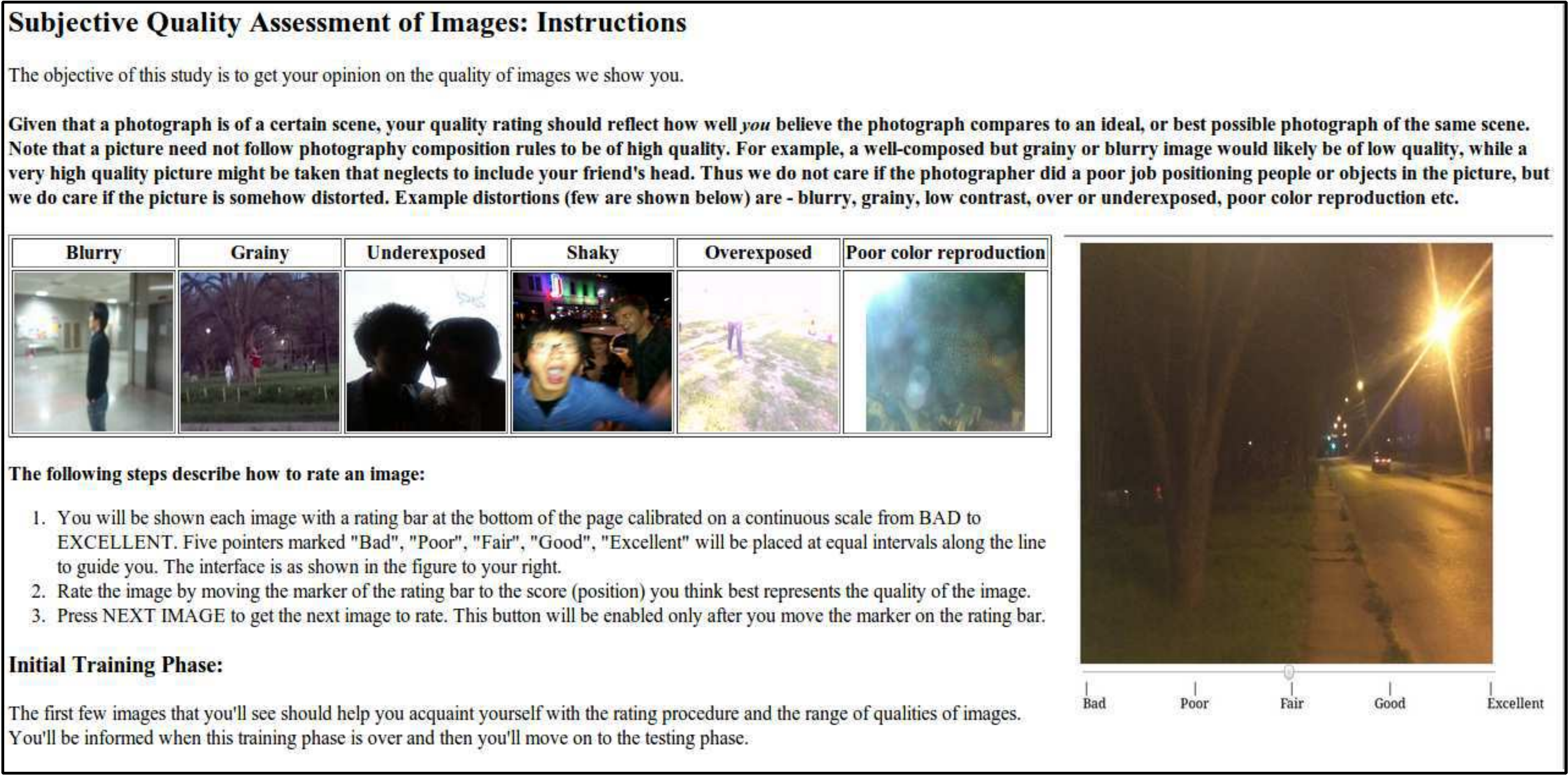}
\vspace{0.5cm}
\caption{\small{Instructions page shown before the worker accepts the task on AMT.}}
\vspace{0.3cm}
\label{fig:instructions-b}
\end{center}
\end{figure*}

Crowdsourcing has been extensively and successfully used on several object identification tasks \cite{labelme,crowdGame} to gather segmented objects and their labels. However, the task of labeling objects is often more clearly defined and fairly straightforward to perform, by contrast with the more subtle, challenging, and highly subjective task of gathering opinion scores on the perceived quality of images. The generally naive level of experience of the workers with respect to understanding the concept of image quality and their geographical diversity made it important that detailed instructions be provided to assist them in understanding how to undertake the task without biasing their perceptual scores. Thus, every unique participating subject on AMT that selects our HIT was first provided with detailed instructions to help them assimilate the task. A screenshot of this web page is shown in Fig. \ref{fig:instructions-b}. Specifically, after defining the objective of the study, a few sample images were presented which are broadly representative of the kinds of distortions contained in the database, to help draw the attention of the workers to the study and help them understand the task at hand. A screenshot of the rating interface was also given on the instructions page, to better inform the workers of the task and to help them decide if they would like to proceed with it. 


\begin{figure*}[t]
\begin{center}
\includegraphics[width=7.5cm]{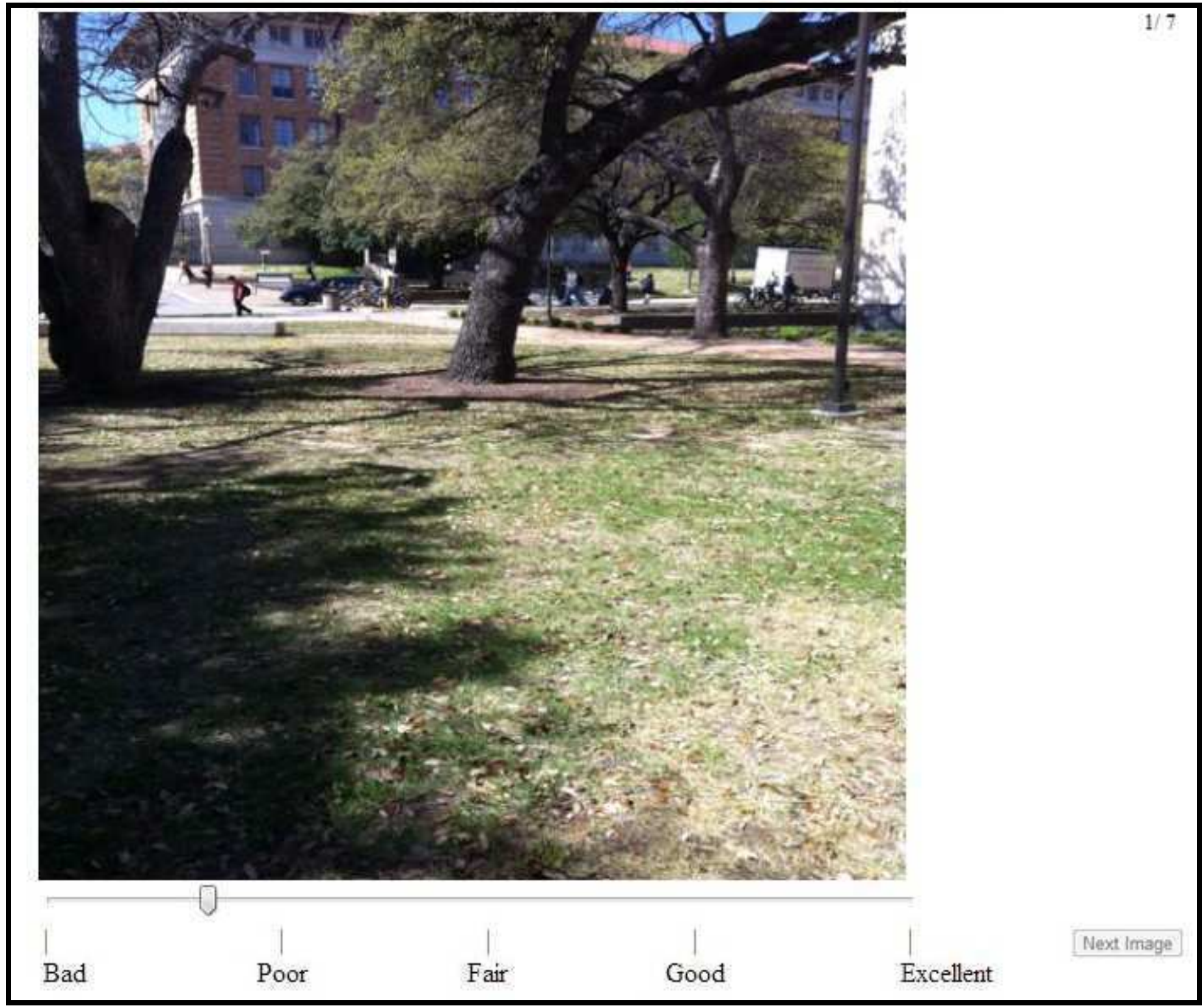}
\caption{\small{The rating interface presented to every subject on which they can provide opinion scores on images.}}
\label{fig:instructions-a}
\end{center}
\end{figure*} 


\textbf{Ensuring unique participants:} After reading the instructions, if a worker accepted the task, and did so for the first time, a rating interface was displayed that contains a slider by which opinion scores could be interactively provided. A screenshot of this interface is also shown in Fig. \ref{fig:instructions-a}. In the event that this worker had already picked our task earlier, we informed the worker that we are in need of unique participants and this worker was not allowed to proceed beyond the instructions page. Only workers with a confidence value\footnote{AMT assigns a confidence score in the range of 0-1 to each worker, based on the accuracy of their responses across all the HITs they have accepted thus far. The higher this number, the more trustworthy a worker is.} greater than 0.75 were allowed to participate. Even with such stringent subject criteria, we gathered more than 350,000 ratings overall. 

\begin{figure*}[t]
\begin{center}
\includegraphics[width=6.1in, height=1.3in]{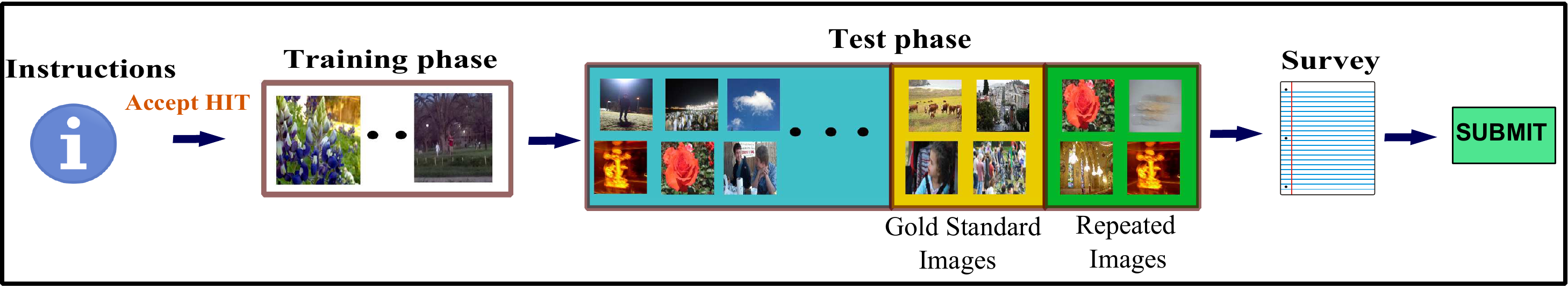}
\caption{\small{Illustrating the design of our HIT. Once a worker clicked the ``Accept HIT'' button and did so for the first time, we directed her to the training phase which was followed by a test phase. A worker who had already participated once in our study and attempted to participate again was not allowed to proceed beyond the instructions page. For the purpose of illustration, we show gold standard and repeated images in exclusion. In reality, the pool of 43 test images was presented in a random order.}}
\label{fig:hit}
\end{center}
\end{figure*}

\begin{figure*}[t]
\begin{center}
\includegraphics[width=9cm]{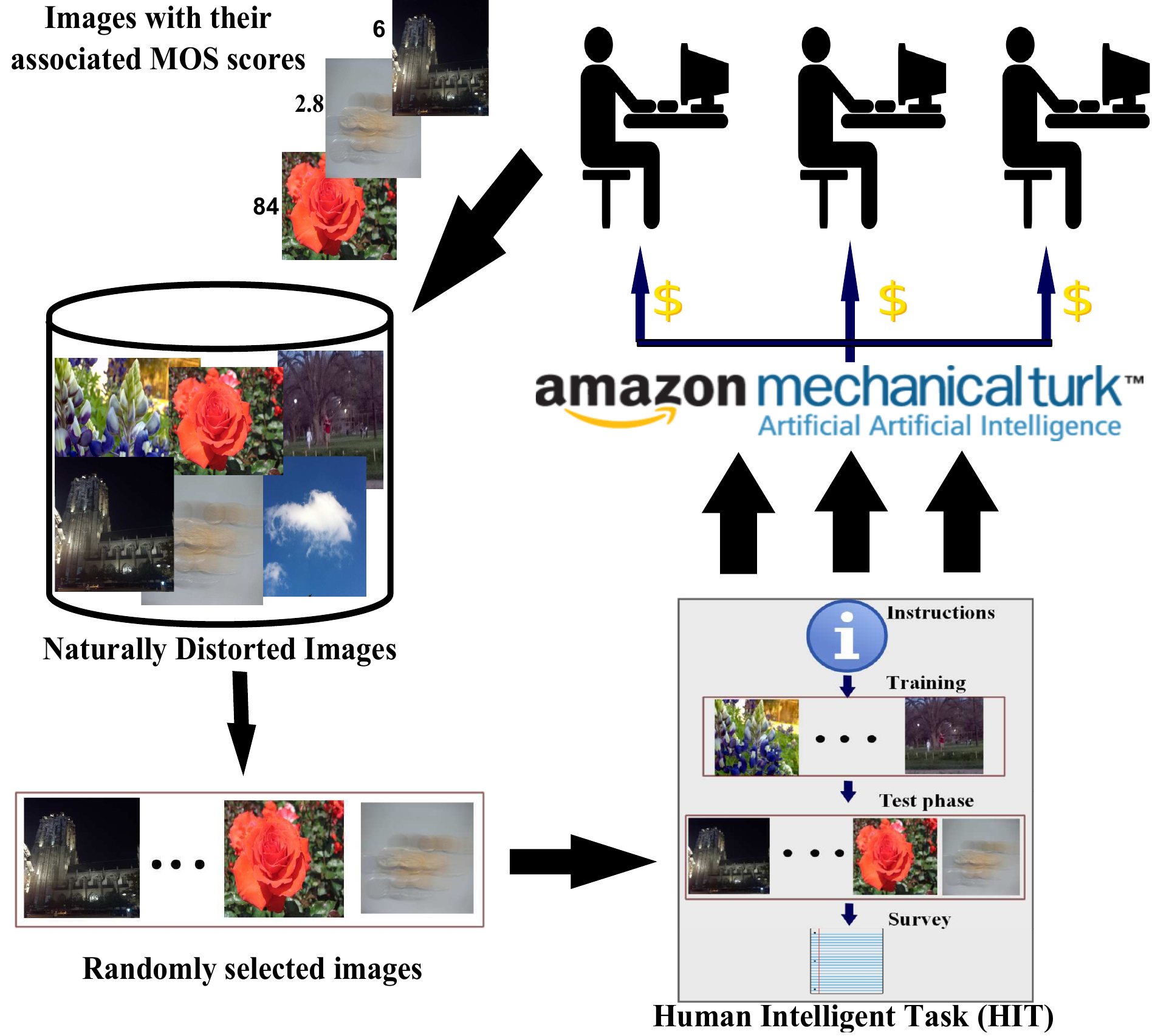}
\caption{\small{Illustrating how our system packages the task of rating images as a HIT and disperses it on Mechanical Turk.}}
\label{fig:turkFlow}
\end{center}
\end{figure*}

\textbf{Study framework:} We adopted a single stimulus continuous procedure \cite{pinson} to obtain quality ratings on images where subjects reported their quality judgments by dragging the slider located below the image on the rating interface. This continuous rating bar is divided into five equal portions, which are labeled ``bad,'' ``poor,'' ``fair,'' ``good,'' and ``excellent.'' After the subject moved the slider to rate an image and pressed the \emph{Next Image} button, the position of the slider was converted to an integer quality score in the range $1-100$, then the next image was presented. Before the actual study began, each participant is first presented with $7$ images that were selected by us as being reasonably representative of the approximate range of image qualities and distortion types that might be encountered. We call this the \textbf{training phase}. Next, in the \textbf{testing phase}, the subject is presented with $43$ images in a random order where the randomization is different for each subject. This is followed by a quick survey session which involves the subject answering a few questions. Thus, each HIT involves rating a total of 50 images and the subject receives a remuneration of $30$ cents for the task. Figure \ref{fig:hit} illustrates the detailed design of our HIT on IQA and Fig. \ref{fig:turkFlow} illustrates how we package the task of rating images as a HIT and effectively disperse it online via AMT to gather thousands of human opinion scores. 
\subsection{Subject Reliability and Rejection Strategies}
Crowdsourcing has empowered us to efficiently collect large amounts of ratings. However, it raises interesting issues such as dealing with noisy ratings and addressing the reliability of the AMT workers. 
\subsubsection{Intrinsic metric}
To gather high quality ratings, only those workers on AMT with a confidence value greater than 75\% were allowed to select our task. Also, in order to not bias the ratings due to a single worker picking our HIT multiple times, we imposed a restriction that each worker could select our task no more than once.

\subsubsection{Repeated images} 5 of each group of 43 test images were randomly presented twice to each subject in the testing phase. If the difference between the two ratings that a subject provided to the same image each time it was presented exceeded a threshold on at least 3 of the 5 images, then that subject was rejected. This served to eliminate workers that were providing unreliable, ``random'' scores. Prior to the full-fledged study, we conducted an initial subjective study and obtained ratings from 300 unique workers. We then computed the average standard deviation of these ratings on all the images. Rounding this value to the closest integer yielded 20 which we then used as our threshold for subject rejection.

\subsubsection{Gold Standard Data} \label{goldStandard}
5 of the remaining 38 test images were drawn from the LIVE Multiply Distorted Image Quality Database \cite{multiply} to supply a control. These images along with their corresponding MOS from that database were treated as a \emph{gold standard}. The mean of the Spearman's rank ordered correlation values computed between the MOS obtained from the workers on the gold standard images and the corresponding ground truth MOS values from the database was found to be \textbf{0.9851}. The mean of the absolute difference between the MOS values obtained from our crowdsourced study and the ground truth MOS values of the gold standard images was found to be \textbf{4.65}. Furthermore, we conducted a paired-sampled t-test and observed that this difference between gold standard and crowdsourced MOS values is not statistically significant. This high degree of agreement between the scores gathered in a traditional laboratory setting and those gathered via an uncontrolled online platform with several noise parameters is critical to us. Although the uncontrolled test settings of an online subjective study could be perceived as a challenge to the authenticity of the obtained opinion scores, this high correlation value indicates a high degree of reliability of the scores that are being collected by us using AMT, reaffirming the efficacy of our approach of gathering opinion scores and the high quality of the obtained subject data.
\subsection{Subject-Consistency Analysis} 
In addition to measuring correlations against the gold standard image data as discussed above, we further analyzed the subjective scores in the following two ways:
\subsubsection{Inter-Subject consistency} To evaluate subject consistency, we split the ratings obtained on an image into two disjoint equal sets, and computed two MOS values on every image, one from each set. When repeated over 25 random splits, an average Spearman's rank ordered correlation between the mean opinion scores between the two sets was found to be \textbf{0.9896}. 
\subsubsection{Intra-Subject consistency} Evaluating intra-subject reliability is a way to understand the degree of consistency of the ratings provided by individual subjects \cite{hossfeld-stall}. We thus measured the Spearman's rank ordered correlation (SROCC) between the individual opinion scores and the MOS values of the gold standard images. A median SROCC of \textbf{0.8721} was obtained over all of the subjects.

All of these additional experiments further highlight the high degree of reliability and consistency of the gathered subjective scores and of our test framework.

\section{Analysis of the Subjective Scores}
\begin{figure}[t]
\begin{center}
\includegraphics[width=4.5cm]{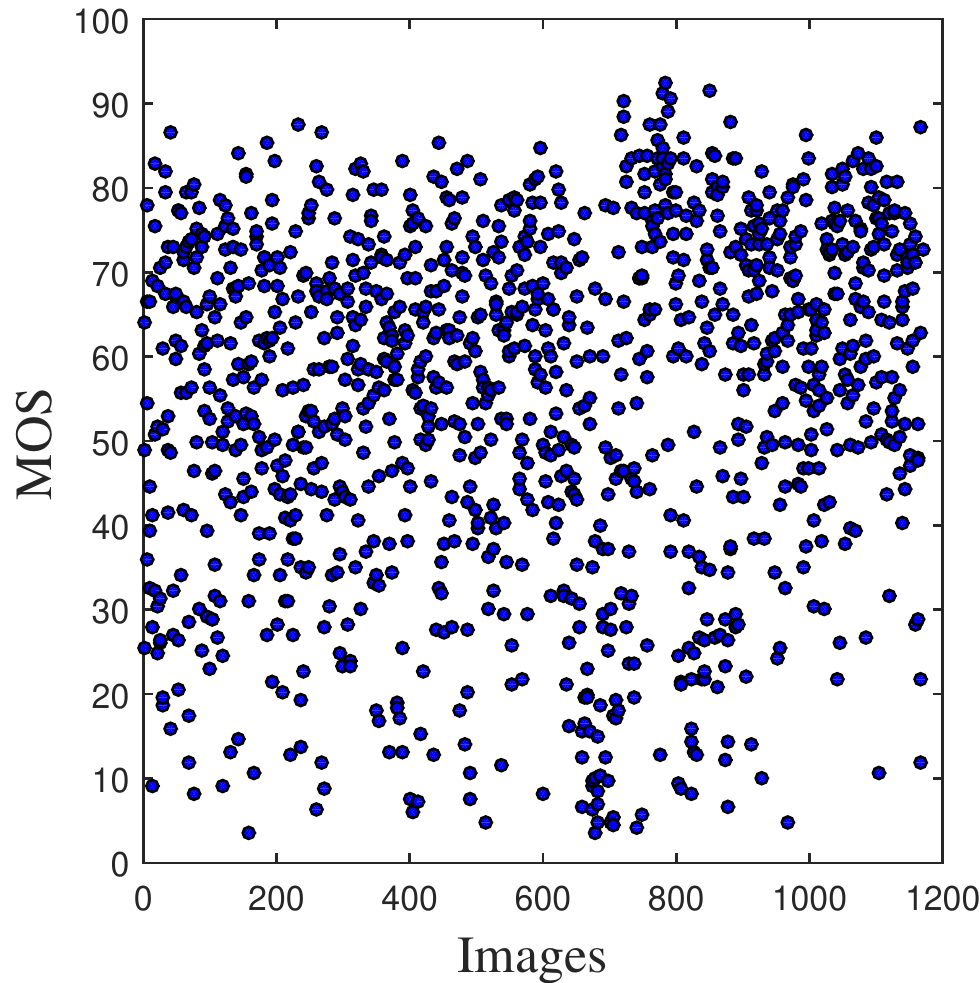}
\caption{\small{Scatter plot of the MOS scores obtained on all the images in the database.}}
\label{fig:MOS}
\end{center}
\end{figure}

The database currently comprises of more than $350,000$ ratings obtained from more than $8,100$ unique subjects (after rejecting unreliable subjects). Enforcing the aforementioned rejection strategies led us to reject 134 participants who had accepted our HIT. Each image was viewed and rated by an average of $175$ unique subjects, while the minimum and maximum number of ratings obtained per image were $137$ and $213$, respectively. While computing these statistics, we excluded the 7 images used in the training phase and the 5 gold standard images as they were viewed and rated by all of the participating subjects. Workers took a median duration of $4.37$ minutes to view and rate all 50 images presented to them. The Mean Opinion Scores (MOS) after subject rejection was computed for each image by averaging the individual opinion scores from multiple workers. MOS is representative of the \emph{perceived viewing experience} of each image. The MOS values range between $[3.42 - 92.43]$. Figure \ref{fig:MOS} is a scatter plot of the MOS computed from the individual scores we have collected. In order to compare the MOS values with single opinion scores (SOS), we computed the standard deviation of the subjective scores obtained on every image and obtained an average standard deviation of $19.2721$. 

\begin{figure}[t] 
\begin{center}$
\begin{array}{cc}
\includegraphics[width=1.6in,height=1.9in]{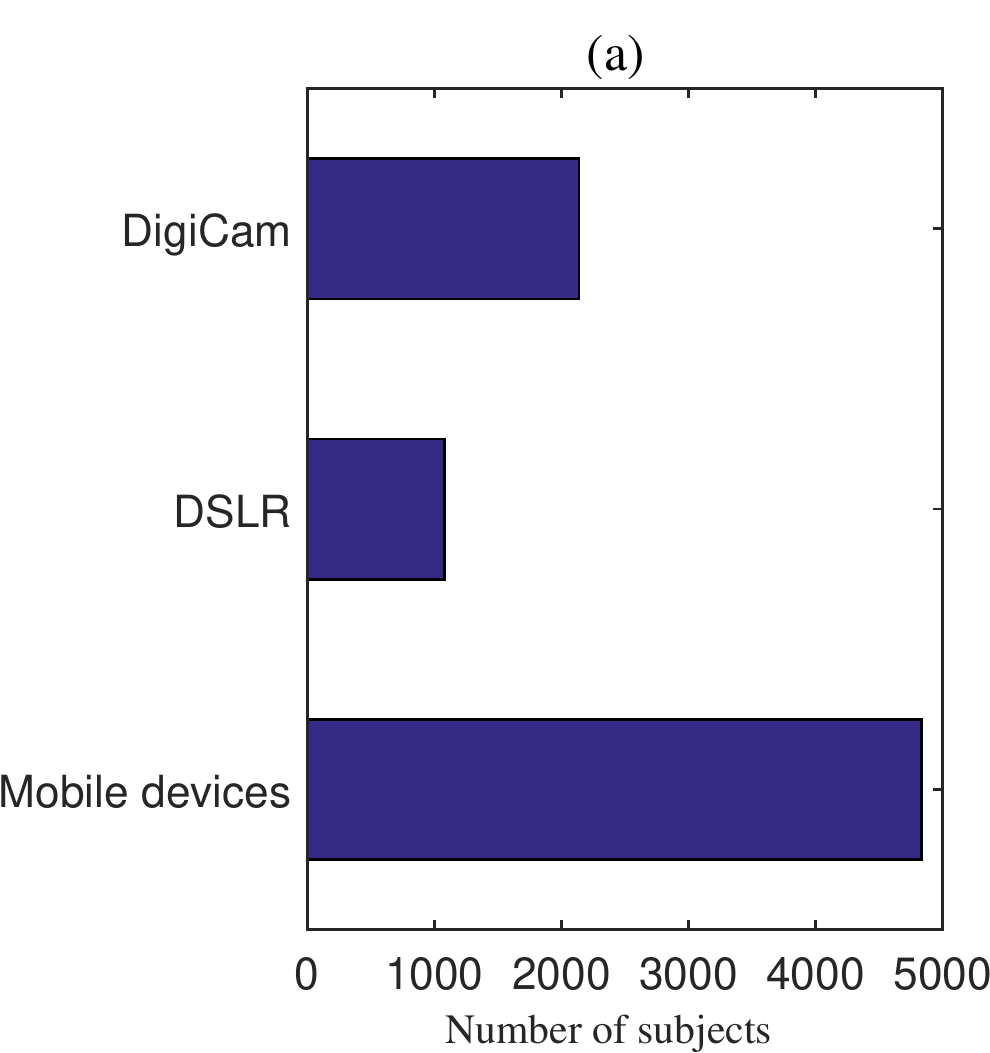} &
\includegraphics[width=1.6in,height=1.9in]{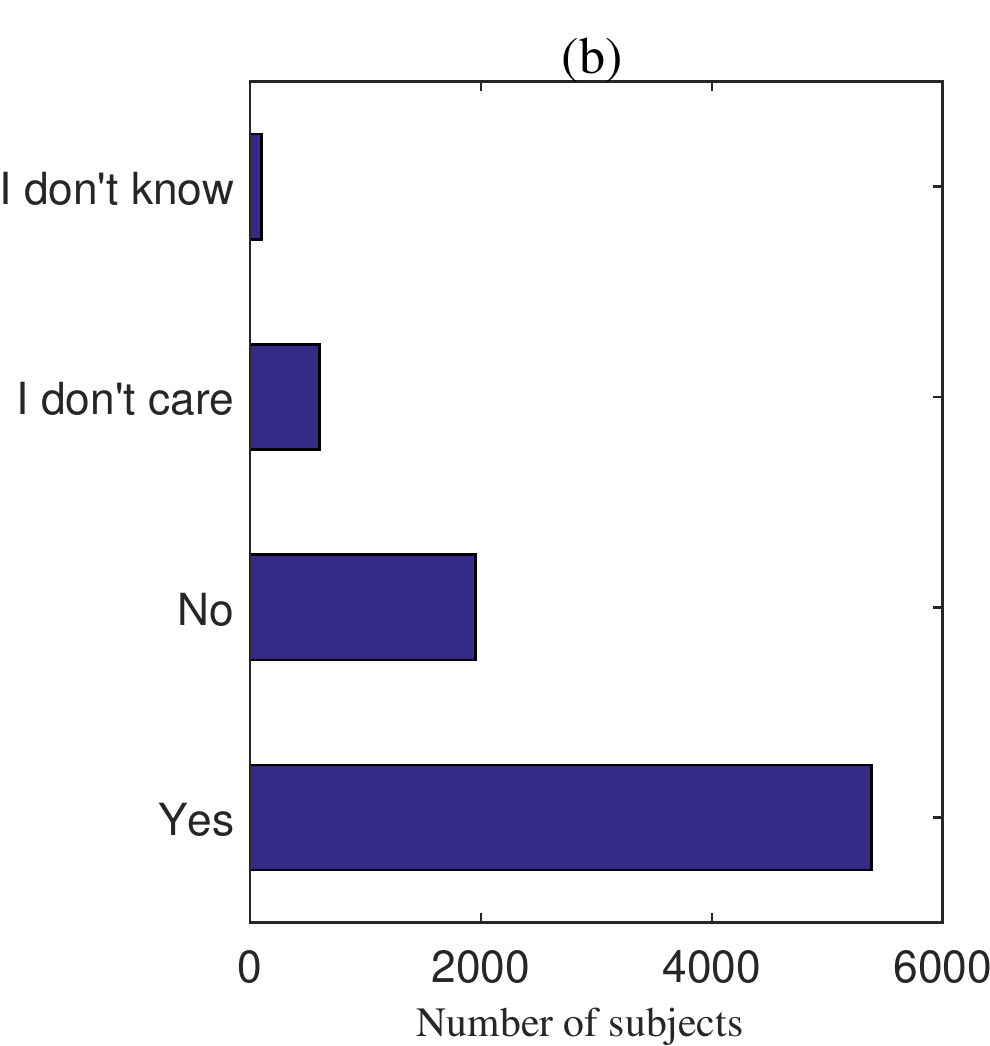}
\end{array}$ 
\caption{\small{Illustrating (a) the kind of consumer image capturing devices preferred by users and (b) their sensitivity to perceived distortions in digital pictures viewed on the Internet.}}
\label{fig:influences1}
\end{center}
\end{figure}

\begin{figure*}[t] 
\begin{center}$
\begin{array}{ccccccc}
\includegraphics[width=1.36in,height=2.0in]{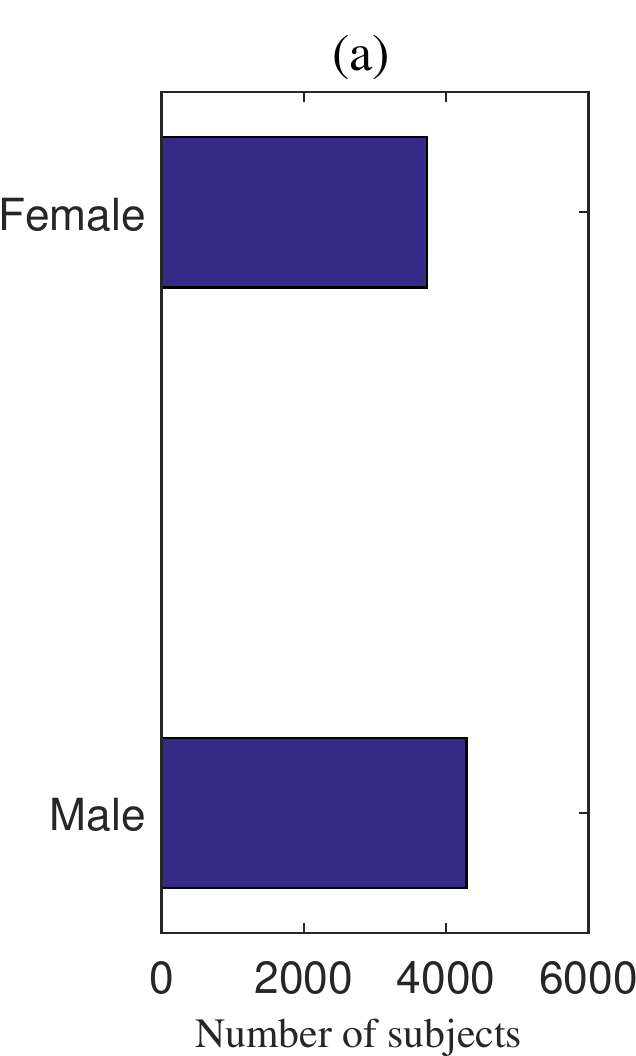} &
\includegraphics[width=1.8in,height=2.0in]{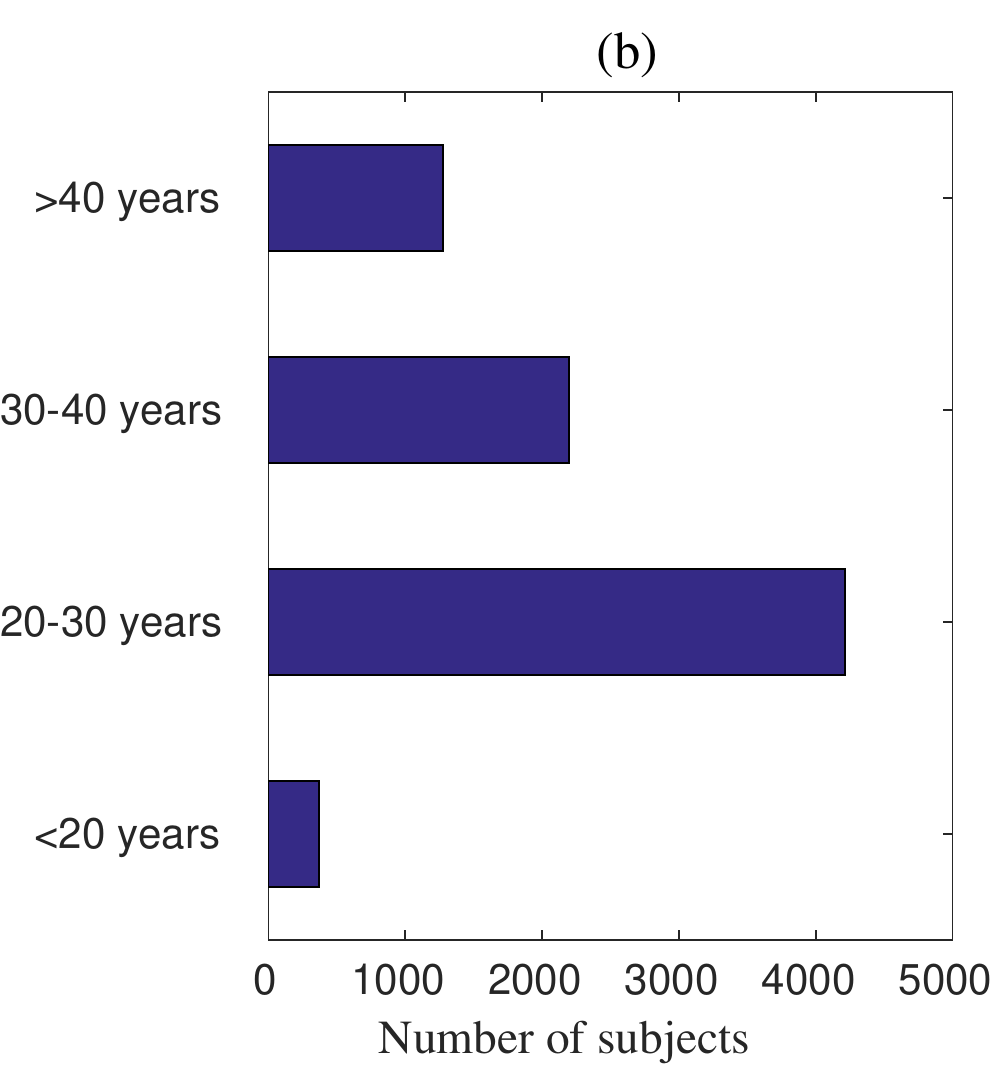} &
\includegraphics[width=1.8in,height=2.0in]{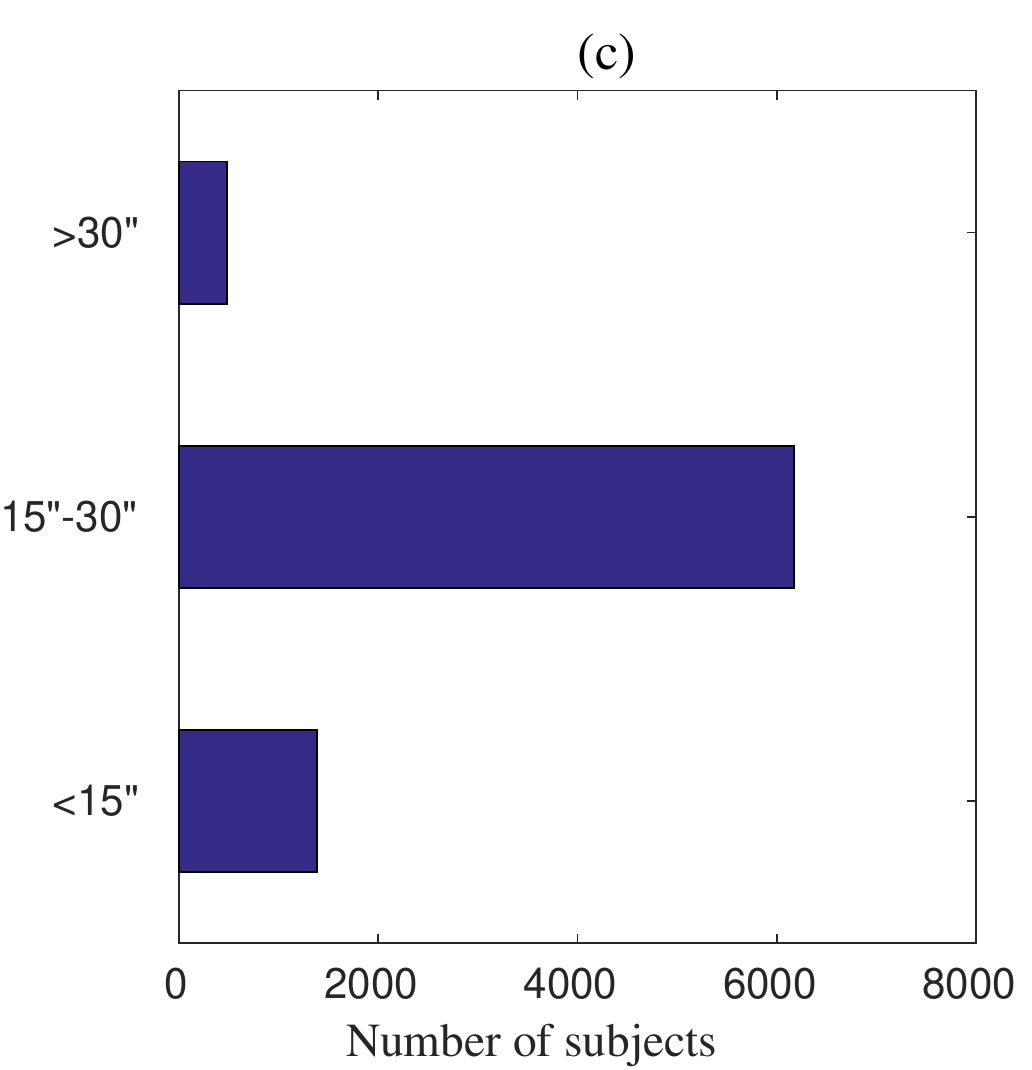} &
\includegraphics[width=1.8in,height=1.99in]{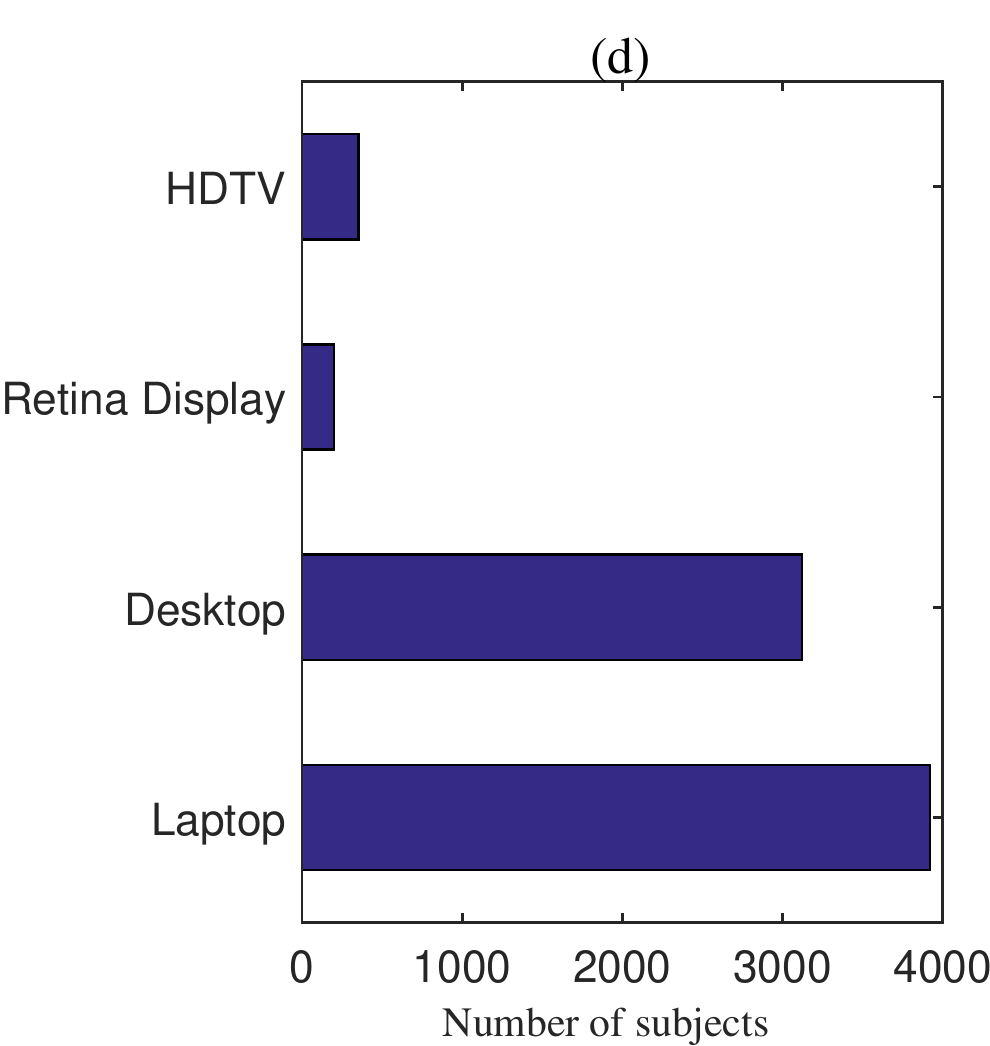}
\end{array}$ 
\caption{\small{Demographics of the participants (a) gender (b) age (c) approximate distance between the subject and the viewing screen (d) different categories of display devices used by the workers to participate in the study.}}
\label{fig:demographics}
\end{center}
\end{figure*}

The uncontrolled online test environment poses certain unique challenges: a test subject of any gender or age may be viewing the image content on any kind of a display, under any sort of lighting, from an unknown distance, and an unknown level of concentration, each of which can affect her choice of quality score. Figures \ref{fig:demographics} (a) and \ref{fig:demographics} (b) illustrate the demographic details of the unique subjects who have participated in our study\footnote{Gathering demographic details of the workers is a common practice on Mechanical Turk. None of the workers expressed any concerns when providing us with these details.}. Most of them reported in the final survey that they are inexperienced with image quality assessment but do get annoyed by image impairments they come across on the Internet. Since we did not test the subjects for vision problems, they were instructed to wear corrective lenses during the study if they do so in their day-to-day life. Later in the survey, the subjects were asked if they usually wore corrective lenses and whether they wore the lenses while participating in the study. The ratings given by those subjects who were not wearing their corrective lenses they were otherwise supposed to wear were rejected. Figures \ref{fig:demographics} (c) and \ref{fig:demographics} (d) illustrate the distribution of the distances from which workers have viewed the images and the broad classes of different display devices used by them. These four plots illustrate the highly varied testing conditions that exist during the online study and also highlight the diversity of the subjects. Figure \ref{fig:influences1} (a) illustrates the distribution of the types of consumer image capture devices that are preferred by the users. It is evident from this plot that most of the workers reported that they prefer using mobile devices to capture photographs in their daily use. One of the questions we posed to our subjects in the survey was whether the poor quality of pictures that they encounter on the Internet bothers them. Subjects chose between the following four options - ``Yes,'' ``No,'' ``I don't really care,'' and ``I don't know.'' The distribution of the responses to this question is plotted in Fig.\ref{fig:influences1} (b) which clearly indicates that a large population of the workers are bothered by poor quality Internet pictures.

We next present our analysis of the influence of several factors such as age, gender, and display devices on user's perceptual quality. In all cases, we study the effect of each factor independently while fixing the values of the rest of the factors. We believe this strategy helped us to closely study the influence of each factor independently and to help avoid combined effects caused by the interplay of several factors on a user's perceptual quality. Note that the results presented in the following sections are consistent irrespective of the specific values that were fixed for the factors.

\begin{figure*}[t] 
\begin{center}$
\begin{array}{ccccc}
\includegraphics[width=1.3in]{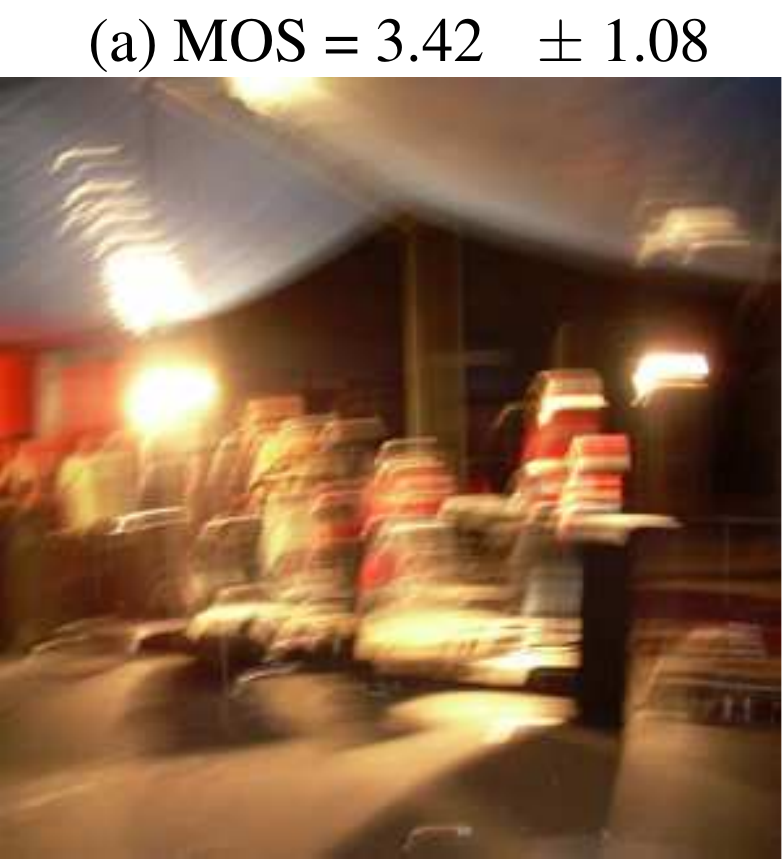} &
\includegraphics[width=1.3in]{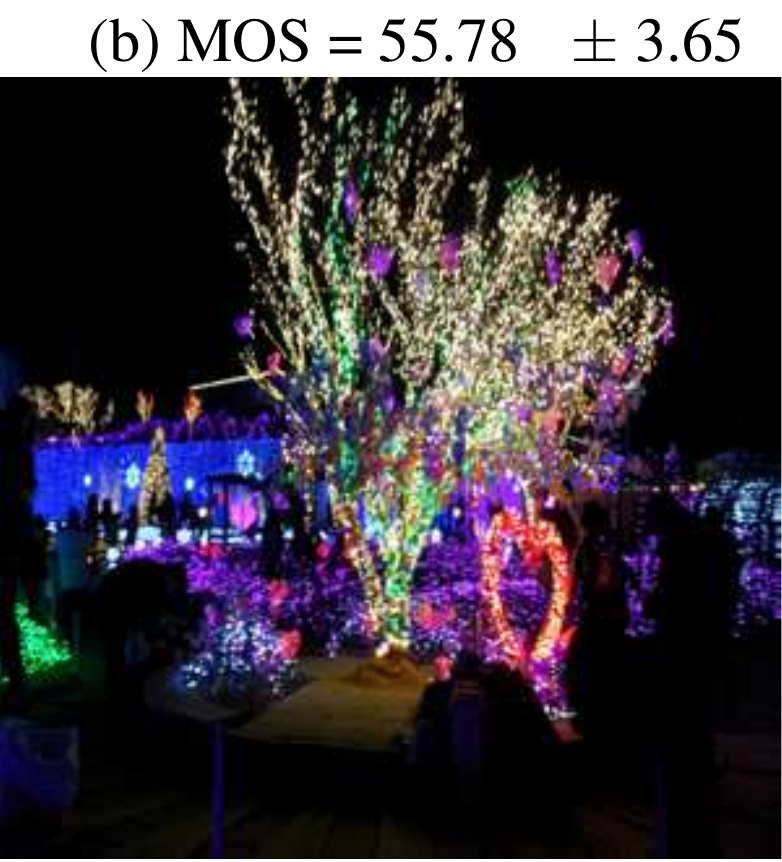} &
\includegraphics[width=1.3in]{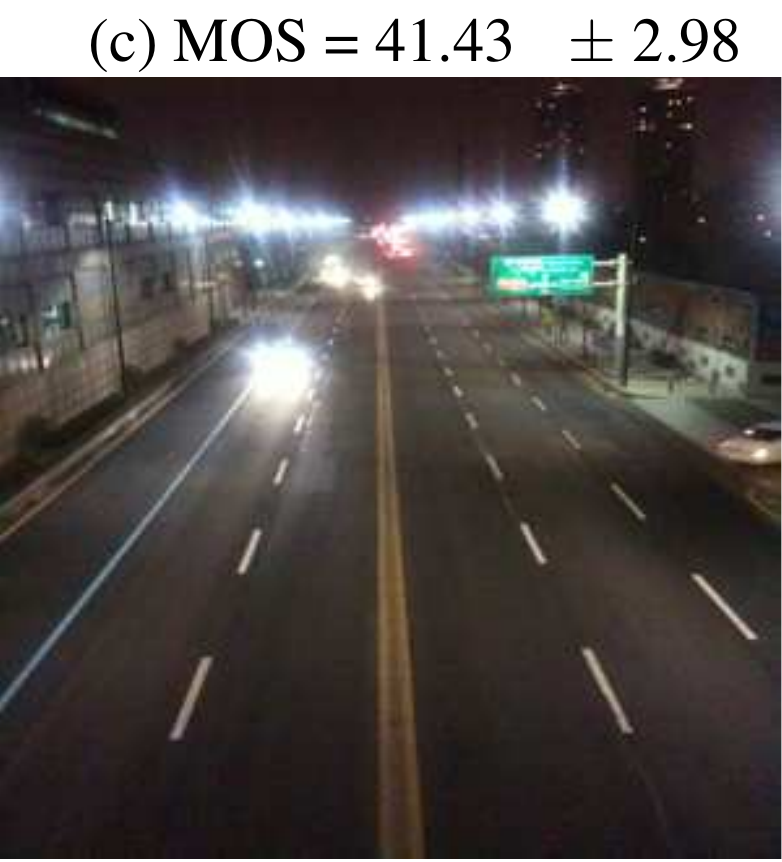} &
\includegraphics[width=1.3in]{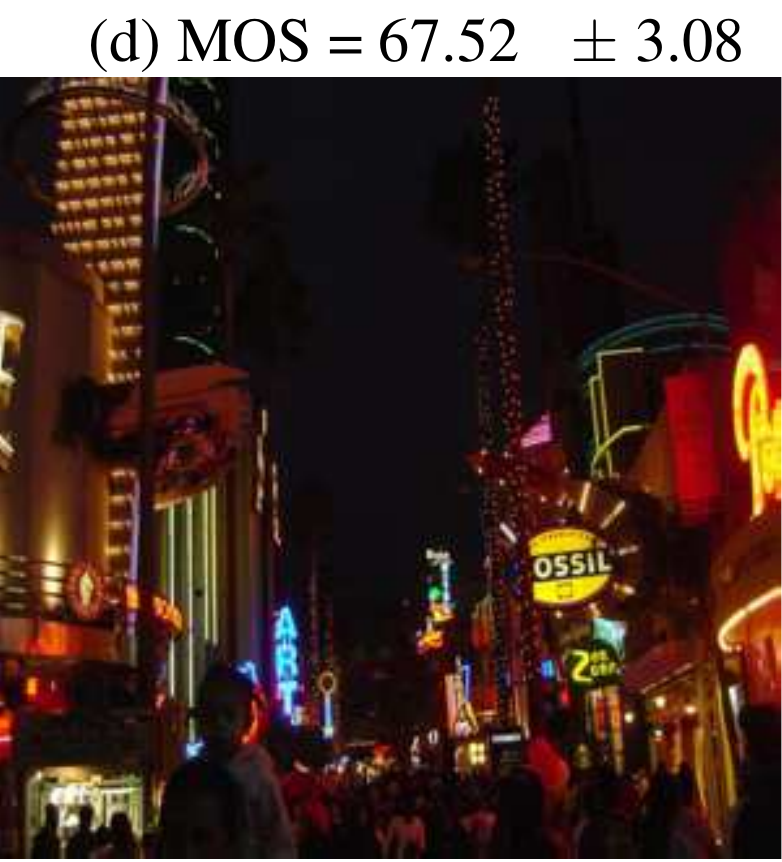} &
\includegraphics[width=1.3in]{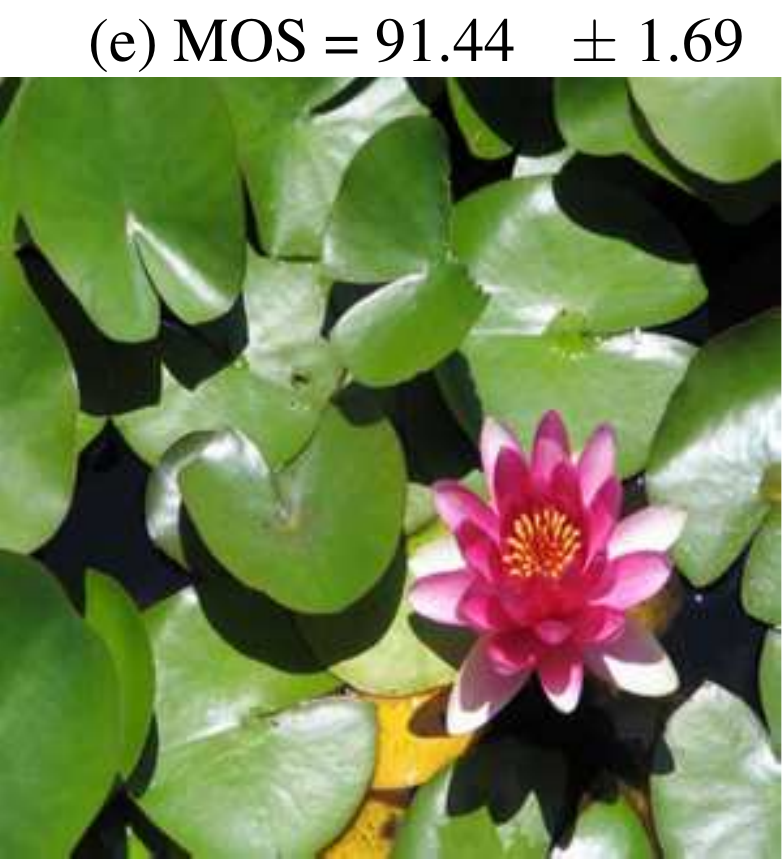} 
\end{array}$
\caption{\small{A few randomly chosen images from the LIVE In the Wild Image Quality Challenge database that are used to illustrate the influence of various parameters on the QoE of the study participants. The upper caption of each image gives the image MOS values and the associated 95\% confidence intervals.}}
\label{fig:infImages}
\end{center}
\end{figure*}

\subsection{Gender}
To understand to what extent gender had an affect on our quality scores, we separately analyzed the ratings obtained from male and female workers on five randomly chosen images (Figures \ref{fig:infImages}(a)-(e)) while maintaining all the other factors constant. Specifically, we separately captured the opinion scores of male and female subjects who are between $20-30$ years old, and reported in our survey to be using a desktop and sitting about $15-30$ inches from the screen. Under this setting and on the chosen set of images, both male and female workers appeared to have rated the images in a similar manner. This is illustrated in Figure \ref{fig:influences}(a).

\subsection{Age} 
\begin{figure*}[t] 
\begin{center}$
\begin{array}{cc}
\includegraphics[width=2.1in,height=2.3in]{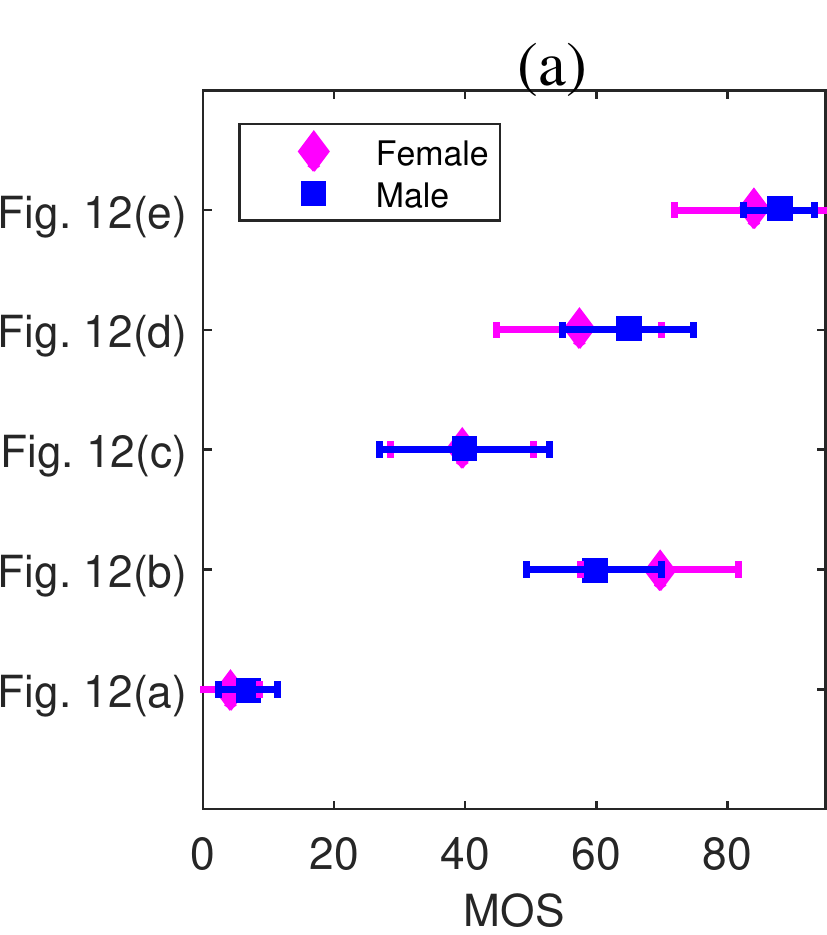} &
\includegraphics[width=2.1in,height=2.3in]{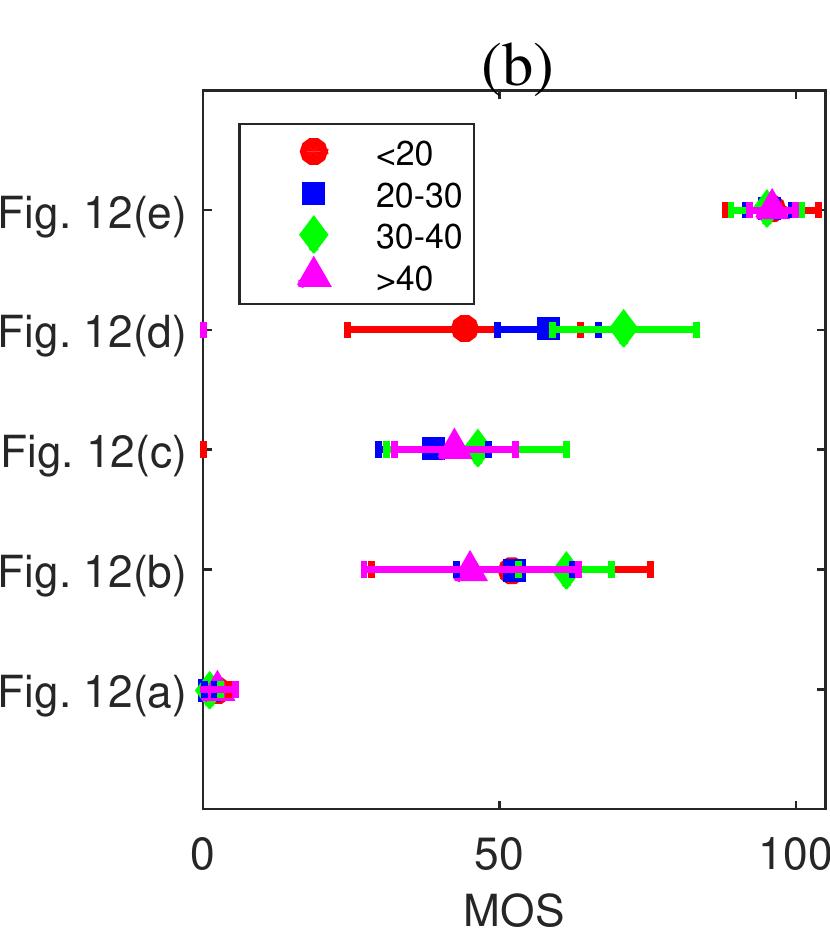} \\
\includegraphics[width=2.2in,height=2.3in]{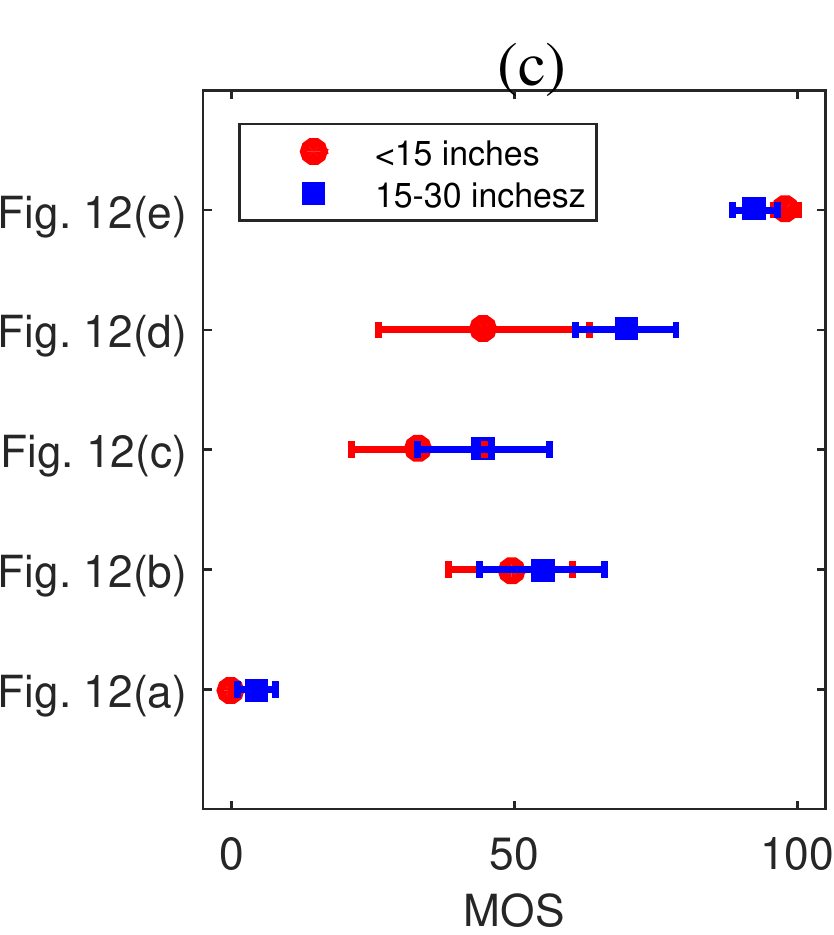} &
\includegraphics[width=2.2in,height=2.3in]{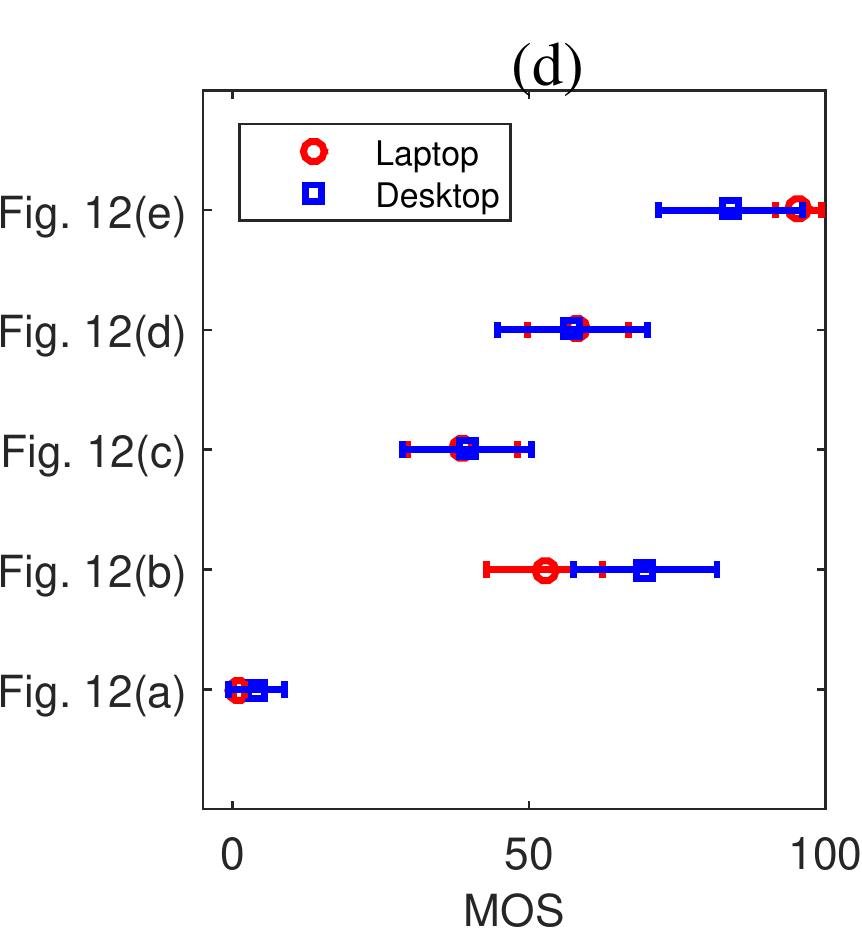}
\end{array}$ 
\caption{\small{Plots showing the influence of a variety of factors on a user's perception of picture quality. The factors are: (a) gender (b) age (c) approximate distance between the subject and the viewing screen and (d) types of display devices used by the workers to participate in the study.The plots detail the range of obtained MOS values and the associated 95\% confidence intervals.}}
\label{fig:influences}
\end{center}
\end{figure*}

Next, we considered both male and female workers who reported using a laptop during the study and were sitting about $15-30$ inches away from their display screen. We grouped their individual ratings on these 5 images (Fig. \ref{fig:infImages}) according to their age and computed the MOS of each group and plotted them in Fig \ref{fig:influences}(b). For the images under consideration, again, subjects belonging to different \emph{age categories} appeared to have rated them in a similar manner. 

Although gender and age did not seem to significantly affect the ratings gathered on the randomly chosen images discussed above, we believe that other factors such as the content in the image can play a significant role in being appealing to one group more than to another. A systematic study focussed exclusively on understanding the interplay of image content, gender, and age using this database might help us better understand the impact of each of these factors on perceptual quality.

\subsection{Distance from the Screen} We next explored the influence of the distance between a subject and her monitor, on the perception of quality. One of the questions in the survey asked the subjects to report which of the three \emph{distance categories} best described a subject's location relative to the viewing screen - ``less than 15 inches,'' ``between 15 to 30 inches,'' and ``greater than 30 inches.''

We gathered the ratings of subjects who reported to be between $30-40$ years old and were participating in the study using their desktop computer. We grouped their ratings\footnote{We received very few ratings from subjects who reported to be sitting greater than 30 inches away from their display screen and hence excluded those ratings from this segment of analysis.} on the five test images (Fig. \ref{fig:infImages}) according to these distance categories and report the results in Fig. \ref{fig:influences}(c). It may be noticed that the difference between the mean of the ratings obtained on the same image when viewed from a closer distance as compared to when the same image was viewed by subjects from a greater distance is not statistically significant. However, we do not rule out the possible influences that viewing distance may have on distortion perception from an analysis of 5 random images. The observed indifference to viewing distance could be due to an interplay of the resolution of the display devices, image content, and viewing distances which is a broad topic worthy of future study.

\subsection{Display Device} To better understand the influence of display devices on QoE, we focussed on workers between $20-30$ years old and who reported to be $15-30$ inches away from the screen while participating in the study. We grouped the ratings of these subjects on the five images in Fig. \ref{fig:infImages} according to the display device that the subjects reported to have used while participating in the study. 

As illustrated in Fig. \ref{fig:influences}(d), the influence of the specific display device that was used for the study appears to have had little effect on the recorded subjective ratings. Of course, we are not suggesting that the perceptual quality of images is unaffected by the display devices on which they are viewed. It is possible that more fine-grained detail regarding the type of display device used by the study participants (e.g., screen resolution, display technology involved, shape of the screen etc.) could deepen our understanding of the dependency between display device and perceptual image quality. However, we chose to focus as much of each participants' effort on the visual tasks as reasonable, and so did not poll them on these details, leaving it for future studies. 
\begin{figure}[t]
\begin{center}
\includegraphics[width=2.1in,height=2.3in]{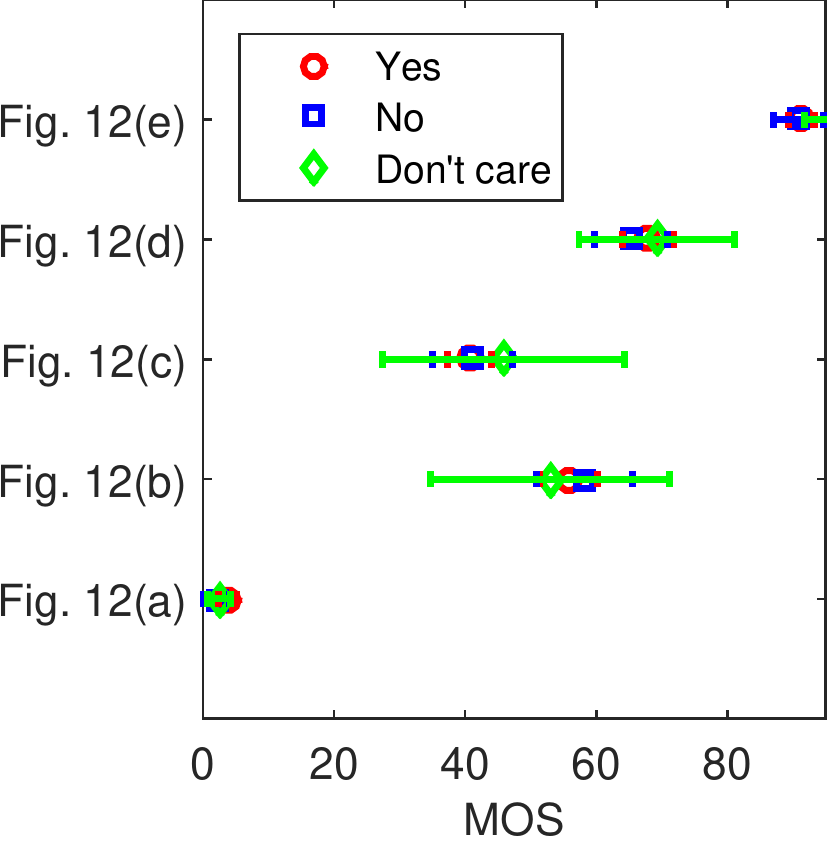}
\caption{\small{Plot showing the influence of users' distortion sensitivity on their quality ratings (along with the associated 95\% confidence intervals.)}}
\label{fig:qualityBothersInf}
\end{center}
\end{figure}
i
\begin{figure}
\begin{center}
\includegraphics[width=3.5in,height=3.3in]{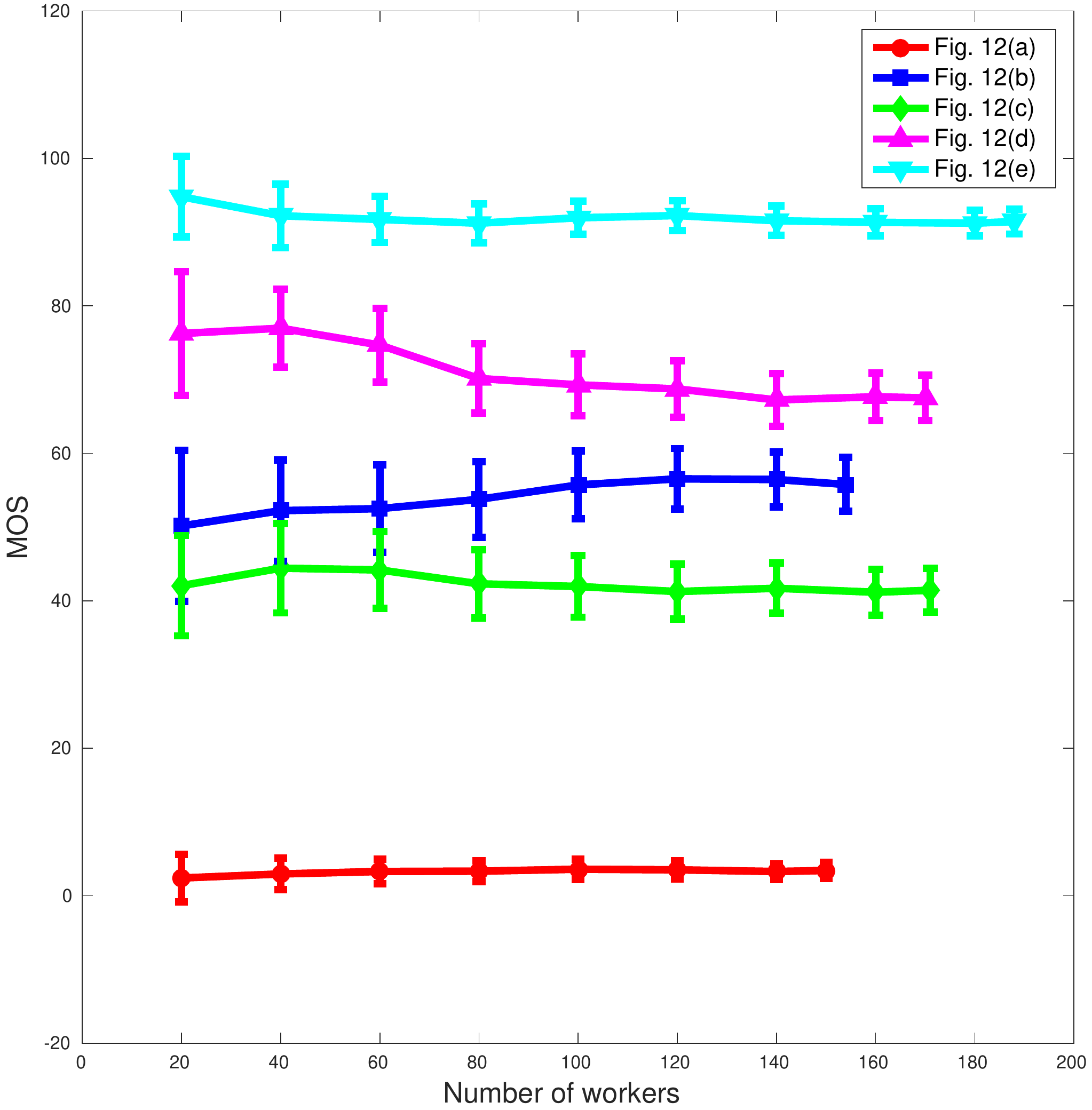} 
\caption{\small{MOS plotted against the number of workers who viewed and rated the images shown in Fig. \ref{fig:infImages}}}
\label{fig:numberOfWorkersInf}
\end{center}
\end{figure}

\subsection{Annoyance of Low Image Quality} As mentioned earlier, one of the questions posed to the subjects in the survey was whether the quality of pictures they encounter on the Internet bothers them (distribution of the responses in Fig. \ref{fig:influences1} (b)). When we grouped our ratings according to these three answers, we noticed that the subjects from each of these three response categories were almost equally sensitive to the visual distortions present in the images from our dataset. This is illustrated in Fig. \ref{fig:qualityBothersInf} (b).

\begin{table*}[t]
\centering
\caption{\small{Summary of our analysis of the different QoE influencing factors on the perception of image distortions.}}
\label{table:infSummary}
\begin{tabular}{|>{\centering\arraybackslash}m{1.3in} | >{\centering\arraybackslash}m{2.1in}  | >{\centering\arraybackslash}m{2.1in} |}
\hline 
{\textbf{Influencing factor}} & {\textbf{Factors that were held constant}} & {\textbf{Observation}} \\
\hline
{Gender} & {Age: 20-30 years, Display device: Desktop, Distance from the screen: 15-30 inches} & {Both male and female workers appeared to rate images in a similar manner.}\\
\hline
{Age} &{Gender: Both male and female workers, Display device: Desktop, Distance from the screen: 15-30 inches}& {Very little difference was noticed in the ratings of the subjects of different age groups.}\\
\hline
{Subject's distance from the display screen} & {Gender: Both male and female workers, Age: 30-40 years, Display device: Desktop} & {Little effect on the subjective ratings.}\\
\hline
{Display device} &{Gender: Both male and female workers, Age: 20-30 years, Distance from the screen: 15-30 inches.}& {Little effect on subjective ratings.}\\
\hline
{Subject's general sensitivity to perceptual quality} & None & {People who claimed to differ in their level of annoyance in response to image distortions appeared to rate images in a similar manner.}\\
\hline
 \end{tabular}
\vspace{0.4cm}
\end{table*} 

Figure \ref{fig:numberOfWorkersInf} illustrates how MOS values flatten out with increases in the number of subjects rating the images. It is interesting to note that there is much more consistency on images with very high and very low MOS values than on intermediate-quality images. Of course, the opinion scores of subjects are affected by several external factors such as the order in which images are presented, a subject's viewing conditions, and so on, and the MOS thus exhibit variability with respect to the number of workers who have rated them.

We summarize all the factors whose influence we studied and presented in this section (by controlling the other factors) in Table \ref{table:infSummary}.

The crowdsourcing image quality study allowed our diverse subjects to participate at their convenience, and in diverse, uncontrolled viewing circumstances, enhancing our ability to investigate the effects of each of these factors on perceived picture quality. The results of our studies of those factors affecting data reliability, and our observations of the high correlations of the objective quality scores against the MOS values of the gold standard images that were obtained under controlled laboratory conditions, both strongly support the efficacy of our online crowdsourcing system for gathering large scale, reliable data. 
\subsection{Limitations of the current study}
Crowdsourcing is a relatively new tool with considerable potential to help in the production of highly valuable and generalized subjective databases representative of human judgments of perceptual quality. However, the approach involves many complexities and potential pitfalls which could affect the veracity of the subject results. A good summary and analysis of these concerns may be found in \cite{hossfeld-stall}.

For example, while we have a high degree of faith in our subject results, it is based on a deep analysis of them rather than simply because the participants were screened to have high AMT confidence values. As mentioned earlier, the confidence values of the workers computed by AMT is an aggregate that is measured over all the tasks in which a worker has participated. This metric thus is not necessarily an indicator of reliability with regards to any specific task and should be accompanied by rigorous, task-specific subject reliability methods. Future studies would benefit by a more detailed data collection and analysis of the details of workers' display devices \cite{hossfeld-stall} and viewing conditions. While our current philosophy, even in laboratory studies, is to not screen the subjects for visual problems, given the newness of the crowdsourcing modality, it might be argued that visual tests could be used to improve subject reliability checks. Many other environmental details could be useful, such as reports of the time spent by a worker in viewing and rating images, to further measure worker reliability.

\section{Experiments}\label{sec:expt}
We also explored the usefulness of the new database by using it to evaluate the quality prediction performance of a number of leading blind IQA algorithms. These algorithms are almost invariably machine learning-based training procedures applied on perceptual and/or statistical image features. Therefore, in all of the experiments described below, we randomly divided the data\footnote{In the results reported in Sec.\ref{sec:diffTech}, $149$ images that were captured during the night were excluded from the dataset. Thus, the total number of images for this experiment consisted of only $1013$ images. In Sec. \ref{sec:nightImages}, the experiments were repeated including these images.} into content-separated disjoint 80\% training and 20\% testing sets, learned a model on the training data, and validated its performance on the test data. To mitigate any bias due to the division of data, we repeated the process of randomly splitting the dataset over 50 iterations and computed Spearman's rank ordered correlation coefficient (SROCC) and Pearson's linear correlation coefficient (PLCC) between the predicted and the ground truth quality scores at the end of every iteration. We report the median of these correlations across 50 iterations. A higher value of each of these metrics indicates better performance both in terms of correlation with human opinion as well as the performance of the model.

\subsection{New Blind Image Quality Assessment Model}
We recently proposed a novel blind image quality assessment model \cite{friquee1, friquee2, globalsip1}, dubbed \textbf{F}eature maps based \textbf{R}eferenceless \textbf{I}mage \textbf{QU}ality \textbf{E}valuation \textbf{E}ngine (FRIQUEE), that seeks to overcome the limitations of existing blind IQA techniques with regards to mixtures of authentic picture distortions, such as those contained in the LIVE In the Wild Image Quality Challenge Database. FRIQUEE is a natural scene statistics (NSS) based model that is founded on the hypothesis that different existing statistical image models capture distinctive aspects of the loss of the perceived quality of a given image. FRIQUEE embodies a total of $564$ statistical features that have been observed to contribute significant information regarding image distortion visibility and perceived image quality. We designed a model that combines a deep belief net and an SVM and achieves superior quality prediction performance when compared to the state-of-the-art. Our proposed deep belief net (DBN) \cite{globalsip1} has four hidden layers formed by stacking multiple restricted boltzmann machines (RBM) and by learning weight matrices at every level by treating the hidden layer's activities of one RBM as the visible input data for training a higher level RBM in a greedy layer-by-layer manner \cite{hintonDigits}. Our DBN builds more complex representations of the simple statistical FRIQUEE features provided as input and remarkably generalizes over different distortion types, mixtures, and severities. The \emph{deep feature representations} learned from our DBN together with subjective opinion scores are later used to train a regressor such that, given a unique test image, its quality is accurately predicted.

\subsection{Comparing the Current IQA Models on the LIVE In the Wild Image Quality Challenge Database}\label{sec:diffTech}
We extracted the quality-predictive features proposed by several prominent blind IQA algorithms (whose code was publicly available) on the images of the LIVE In the Wild Image Quality Challenge Database and used the same learning model that was originally presented in their work\footnote{In the case of DIIVINE \cite{diviine} and C-DIIVINE \cite{cdiivine} which are two-step models, we skipped the first step of identifying the probability of an image belonging to one of the five distortion classes present in the legacy LIVE IQA Database as it doesn't apply to the newly proposed database. Instead, after extracting the features as proposed in their work, we learned a regressor on the training data.} under the 80-20 train-test setting. In this experiment, we excluded $149$ images that were captured during the night. We studied the influence of including night images in the data in Sec. \ref{sec:nightImages}. In the case of FRIQUEE, a combination of DBN and SVM was used and the DBN's input layer has $564$ units, which is equal to the number of features extracted from an image. We report the results in Table \ref{dbn-perf} from which we conclude that the performance of our proposed model on unseen test data is significantly better than current top-performing state-of-the-art methods when tested on the LIVE In the Wild Image Quality Challenge Database.
\begin{table}[t] 
\centering
\caption{\small{Median plcc and Median srocc across 50 train-test combinations on the LIVE In the Wild Image Quality Challenge Database.}}
 \begin{tabular}{| >{\centering\arraybackslash}m{0.9in} | >{\centering\arraybackslash}m{0.3in} | >{\centering\arraybackslash}m{0.3in} |  }
    \hline
&  {PLCC} & {SROCC} \tabularnewline
    \hline
FRIQUEE & \textbf{0.71} & \textbf{0.68} \tabularnewline
\hline
BRISQUE \cite{brisque}&  0.61  & 0.60 \tabularnewline
\hline
DIIVINE \cite{diviine}& 0.56 & 0.51 \tabularnewline
\hline
BLIINDS-II \cite{bliinds2} & 0.45 & 0.40 \tabularnewline
\hline
NIQE \cite{niqe} & 0.48 & 0.42\tabularnewline 
\hline
S3 index \cite{s3Index} & 0.32 & 0.31\tabularnewline
\hline
C-DIIVINE \cite{cdiivine} & 0.66 & 0.63\tabularnewline
\hline
\end{tabular} 
 \label{dbn-perf}
\vspace{0.25cm}
\end{table} 

\subsection{Comparison on a Benchmark Legacy Database}\label{sec:live-r2}
To further highlight the challenges that the authentic distortions present in our database pose to top-performing algorithms, we also computed the median correlation values when the algorithms were tested on the standard legacy benchmark database \cite{live-r2}. FRIQUEE was implemented using the DBN model, by extracting $564$ features on all the images of the dataset and repeating the same evaluation procedure over $50$ iterations. For the other blind IQA models, we use the same learner that was presented originally in their work. We present the median correlation values obtained in Table \ref{legacy-results}. It may be observed that all of the top-performing models, when trained and tested on the legacy LIVE IQA database which is comprised of singly distorted images, perform remarkably well as compared to their performance on our difficult database of images suffering from unknown mixtures of distortions (Table \ref{dbn-perf}). 

\begin{table}[t]
\centering
\caption{\small{Performance on legacy live iqa database \cite{live-r2}. Italics indicate NR IQA models. -NA- indicates data not reported in the corresponding paper.}}
 \begin{tabular}{| >{\centering\arraybackslash}m{1.3in} | >{\centering\arraybackslash}m{0.3in} | >{\centering\arraybackslash}m{0.3in} |  >{\centering\arraybackslash}m{0.3in} | >{\centering\arraybackslash}m{0.3in} | }
\hline
& {SROCC} & {PLCC} \tabularnewline
\hline
PSNR & 0.86 & 0.86\tabularnewline
\hline
SSIM \cite{ssim} & 0.91 & 0.91 \tabularnewline
\hline
MS-SSIM \cite{ms-ssim} & 0.95 & 0.95 \tabularnewline
\hline
\textit{CBIQ} \cite{cbiq} & \textit{0.89} & \textit{0.90} \tabularnewline
\hline
\textit{LBIQ} \cite{lbiq} & \textit{0.91} & \textit{0.91}\tabularnewline
\hline
\textit{DIIVINE} \cite{diviine} & \textit{0.92} & \textit{0.93} \tabularnewline
\hline
\textit{BLIINDS-II} \cite{bliinds2} & \textit{0.91} & \textit{0.92} \tabularnewline
\hline
\textit{BRISQUE} \cite{brisque} & 0.94 & \textit{0.94} \tabularnewline
\hline
\textit{NIQE} \cite{niqe} & 0.91 & 0.92 \tabularnewline
\hline
\textit{C-DIIVINE} \cite{cdiivine} & 0.94 & 0.95 \tabularnewline
\hline
\textit{Tang et.al's model} \cite{tang-cvpr} & \textit{\textbf{0.96}} & \textit{-NA-} \tabularnewline
\hline
\textit{FRIQUEE-ALL} & \textit{0.93} & \textit{0.95} \tabularnewline
\hline
 \end{tabular}
\vspace{0.4cm}
\label{legacy-results}
\end{table}

\begin{table}[t]
\centering
\caption{\small{Median lcc and Median srocc across 100 train-test combinations on the day+night images of LIVE In the Wild Image Quality Challenge Database when svm was used.}}
 \begin{tabular}{| >{\centering\arraybackslash}m{0.8in} | >{\centering\arraybackslash}m{0.3in} | >{\centering\arraybackslash}m{0.3in} |  >{\centering\arraybackslash}m{0.3in} | >{\centering\arraybackslash}m{0.3in} | }
\hline
&  {PLCC} & {SROCC} \tabularnewline
\hline
FRIQUEE & \textbf{0.71} & \textbf{0.70} \tabularnewline
\hline
BRISQUE & 0.60 & 0.60 \tabularnewline
\hline
DIIVINE & 0.60 & 0.59 \tabularnewline
\hline
BLIINDS-II  & 0.44 & 0.46 \tabularnewline
\hline
NIQE & 0.47 & 0.45 \tabularnewline
\hline
C-DIIVINE \cite{cdiivine} & 0.65 & 0.64 \tabularnewline
\hline
 \end{tabular}
\label{dbn-night}
\vspace{0.25cm}
\end{table}

\begin{figure}[t] 
\begin{center}$
\begin{array}{cccc}
\includegraphics[width=0.7in]{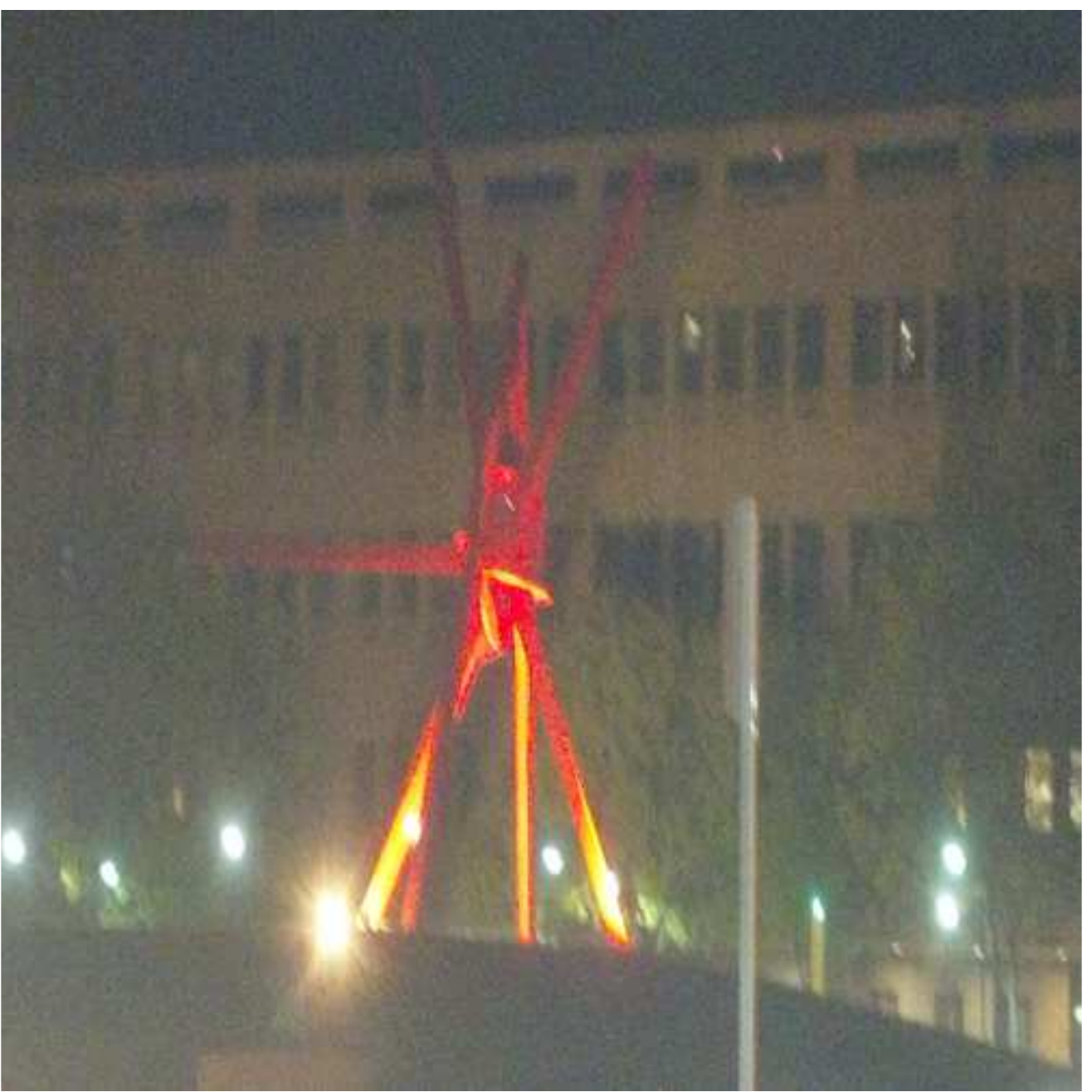} &
\includegraphics[width=0.7in]{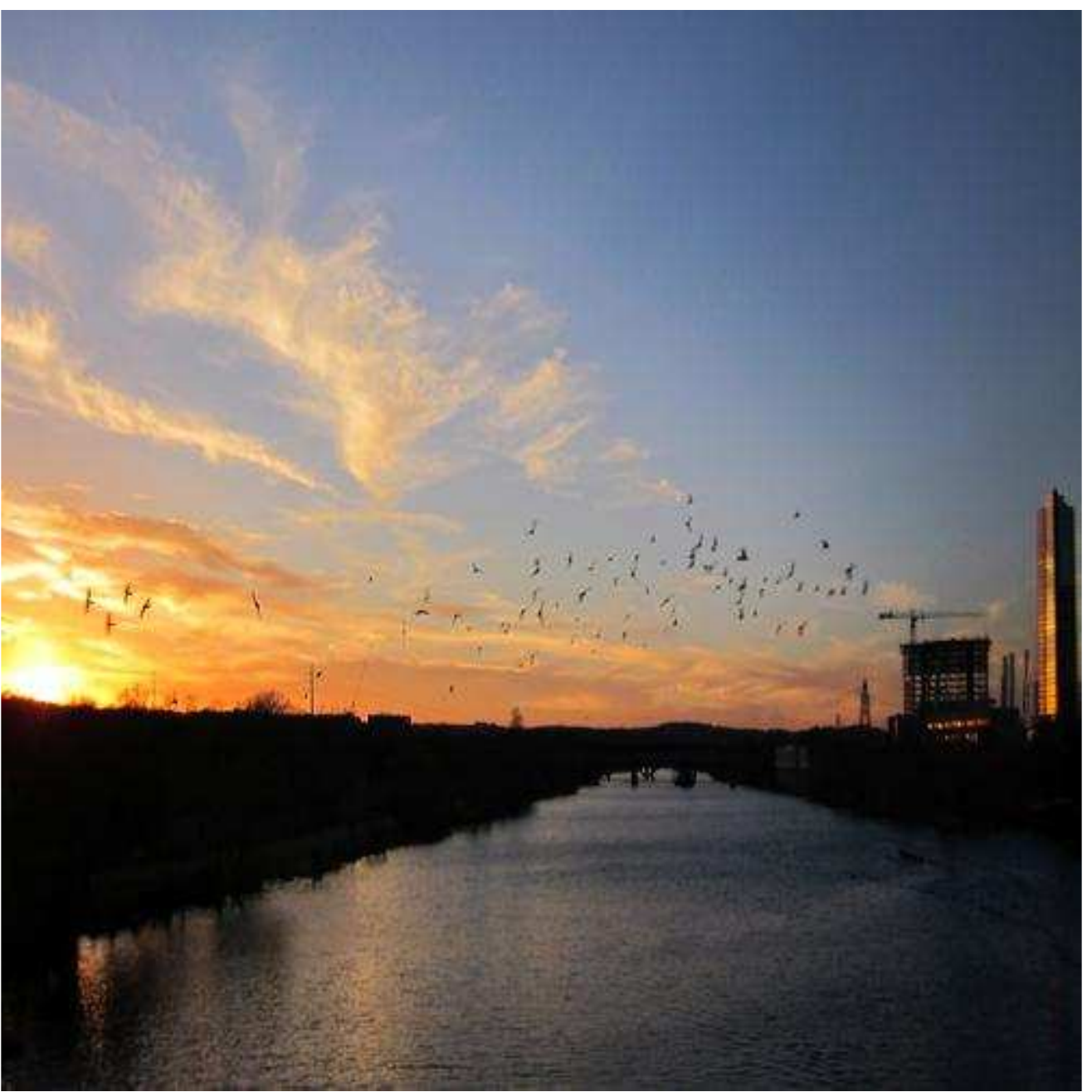} &
\includegraphics[width=0.7in]{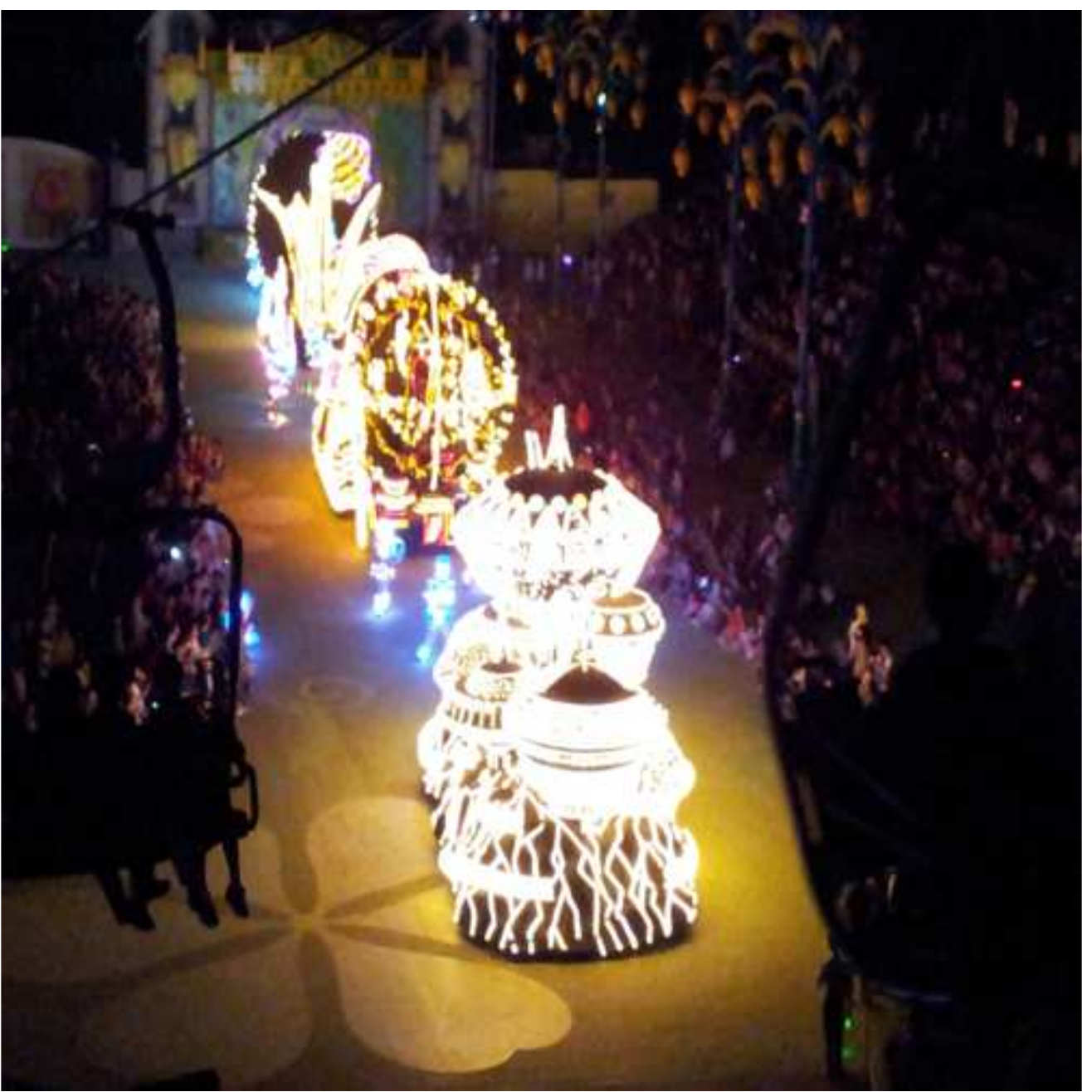} &
\includegraphics[width=0.7in]{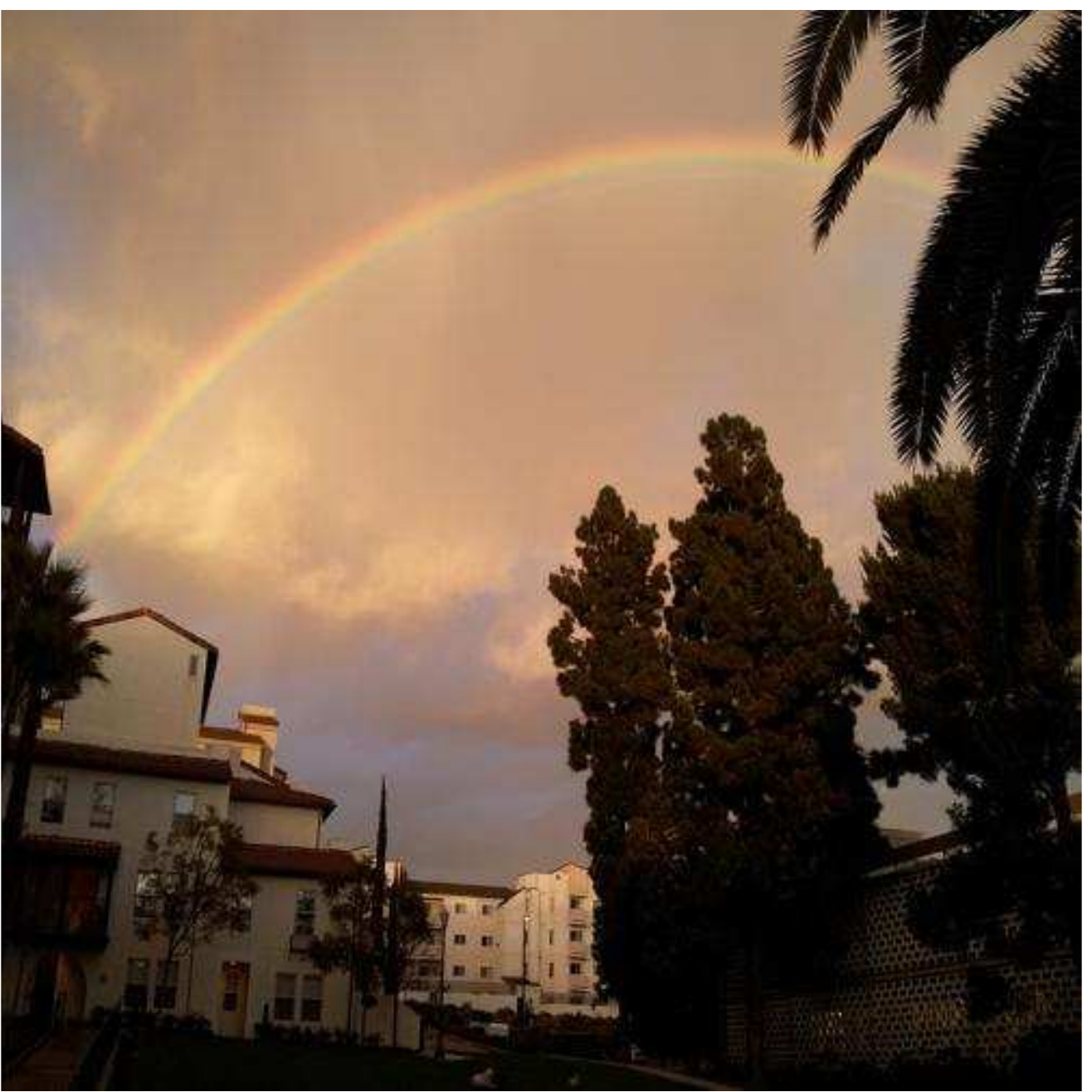} \\
\end{array}$
\caption{\small{Few images from LIVE In the Wild Image Quality Challenge database which were captured during the night.}}
\label{fig:nightImgs}
\end{center}
\end{figure}

\subsection{With and Without Night-time Images}\label{sec:nightImages}
Of the total of $1,162$ images, $149$ pictures were captured at night and suffer from severe low-light distortions\footnote{We interchangbly use `low-light images' or `night images' to refer to these images.} (Fig. \ref{fig:nightImgs}). It should be noted that none of the legacy benchmark databases have images captured under such low illumination conditions and it follows that the NSS-based features used in other models were created by training natural images under normal lighting conditions. Here, we probe the predictive capabilities of top-performing blind IQA models when such low-light images are included in the training and testing data. We therefore included the night-time pictures into the data pool and trained FRIQUEE and the other blind IQA models. The results are given in Table \ref{dbn-night}. Despite such challenging image content, FRIQUEE still performed well in comparison with the other state-of-the-art models. This further supports the idea that a generalizable blind IQA model should be trained over mixtures of complex distortions, and under different lighting conditions. 
\begin{table}[t]
\centering
\caption{\small{Median lcc, Median srocc, Mean Precision (MP) and Mean Recall (MR) across 100 train-test combinations on different combinations of image datasets when DBN proposed in \cite{friquee1, friquee2} was used.}}
 \begin{tabular}{| >{\centering\arraybackslash}m{2.0in} | >{\centering\arraybackslash}m{0.3in} | >{\centering\arraybackslash}m{0.3in} | }
    \hline
Image datasets&  {PLCC} & {SROCC} \tabularnewline
    \hline
LIVE IQA \cite{live-r2}+ LIVE Challenge& 0.82 & 0.78  \tabularnewline
\hline
LIVE Challenge + LIVE Multiply \cite{multiply}& 0.80  & 0.79   \tabularnewline
\hline
LIVE Challenge + LIVE Multiply \cite{multiply} + LIVE IQA \cite{live-r2} & 0.85  & 0.83   \tabularnewline
\hline
\end{tabular}
 \label{tbl:combine-perf}
\end{table}
\subsection{Combining Different IQA Databases}\label{sec:combineDatabases}
Since NR IQA algorithms are generally trained and tested on various splits of a single dataset (as described above), it is natural to wonder if the trained set of parameters are database specific. In order to demonstrate that the training process is simply a calibration, and that once such training is performed, an ideal blind IQA model should be able to assess the quality of any distorted image (from the set of distortions it is trained for), we evaluated the performance of the multi-model FRIQUEE algorithm on combinations of different image databases - the LIVE IQA Database \cite{live-r2} and the LIVE Multiply Distorted IQA Database \cite{multiply}, as well as the new LIVE In the Wild Image Quality Challenge Database. The same 80-20 training setup was followed, i.e., after combining images from the different databases, 80\% of the randomly chosen images were used to train our DBN model and the trained model was then tested on the remaining 20\% of the image data. We present the results in Table \ref{tbl:combine-perf}. It is clear from Table \ref{tbl:combine-perf} that the performance of FRIQUEE is \emph{not} database dependent and that once trained, it is capable of accurately assessing the quality of images across the distortions (both single and multiple, of different severities) that it is trained for. The results clearly show FRIQUEE's potential to tackle the imminent deluge of visual data and the unavoidable distortions they are bound to contain.
\section{Conclusions and Future Work}
With more than $350,000$ subjective judgments overall, we believe that the study described here is the largest, most comprehensive study of perceptual image quality ever conducted. Of course, digital videos (moving pictures) are also being captured with increasing frequency by both professional and casual users. In the increasingly mobile environment, these spatial-temporal signals will be subject to an even larger variety of distortions \cite{bovik13} arising from a multiplicity of natural and artifical processes \cite{bovikBook}. Predicting, monitoring, and controlling the perceptual effects of these distortions will require the development of powerful blind video quality assessment models, such as \cite{saad}, and new VQA databases representative of human opinions of modern, realistic videos captured by current mobile video camera devices and exhibiting contemporary distortions. Current legacy VQA databases, such as \cite{vqa1, vqa2} are useful tools but are limited in regard to content diversity, numbers of subjects, and distortion realism and variability. Therefore, we plan to conduct large-scale crowdsourced \emph{video} quality studies in the future, mirroring the effort described here, and building on our expertise in conducting the current study.

\section{Acknowledgements}
We acknowledge Prof. Sanghoon Lee, Dr. Anish Mittal, Dr. Rajiv Soundararajan, numerous unnamed photographers from UT Austin and Yonsei University, among others, for helping to collect the original images in the LIVE In the Wild Image Quality Challenge Database. This work was supported in part by the National Science Foundation under Grant IIS-1116656.

\begin{IEEEbiography}[{\includegraphics[width=1in,height=1.25in, clip,keepaspectratio]{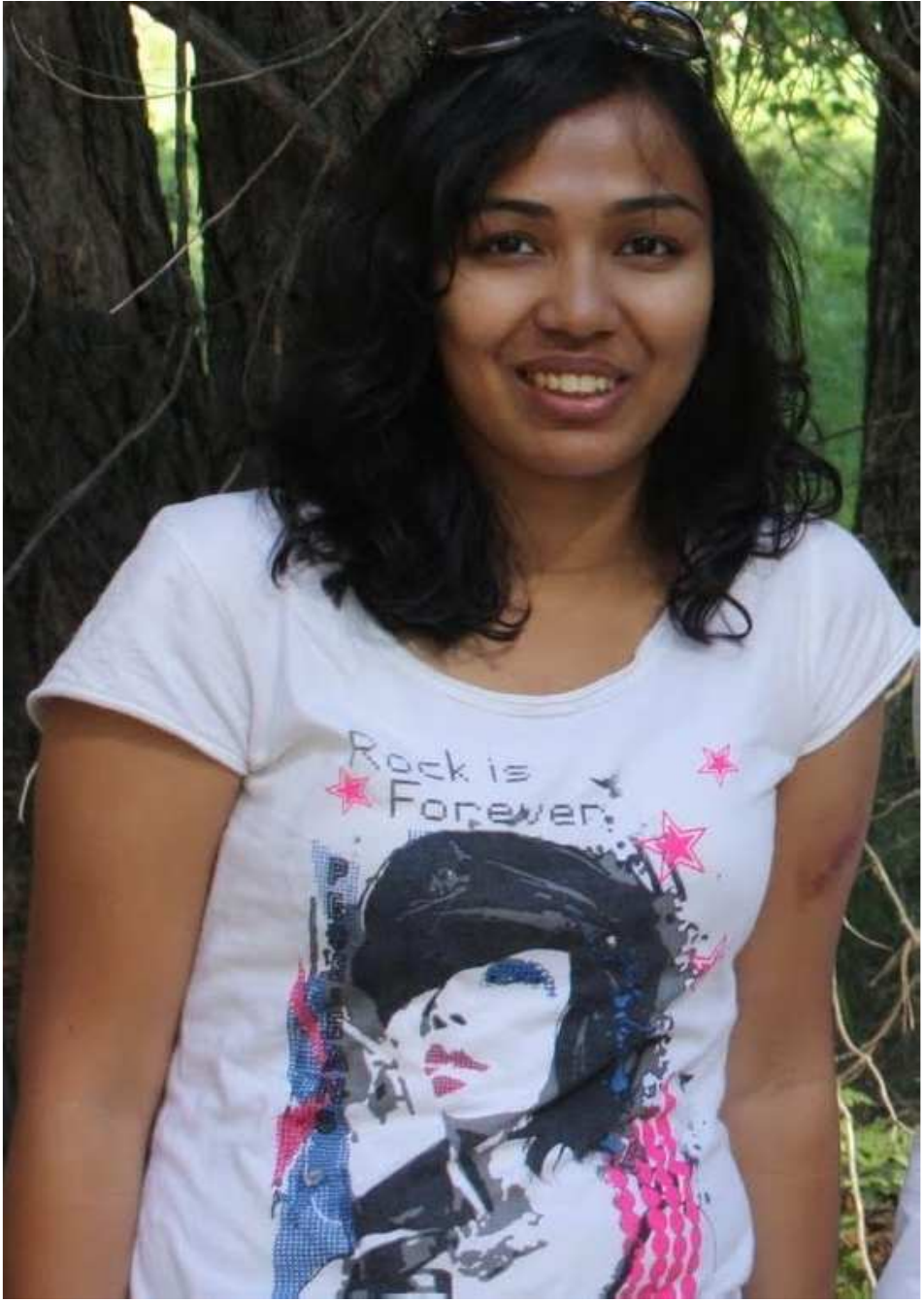}}]{Deepti Ghadiyaram}
received the B.Tech. degree in Computer Science from International Institute of Information Technology (IIIT), Hyderabad in 2009, and the M.S. degree from the University of Texas at Austin in 2013. She joined the Laboratory for Image and Video Engineering (LIVE) in January 2013, and is currently pursuing her Ph.D. She is the recipient of the Microelectronics and Computer Development (MCD) Fellowship, 2013-2014. Her research interests include image and video processing, computer vision, and machine learning, and their applications to the aspects of information retrieval such as search and storage. 
\end{IEEEbiography}

\begin{IEEEbiography}[{\includegraphics[width=1in,height=1.25in,clip,keepaspectratio]{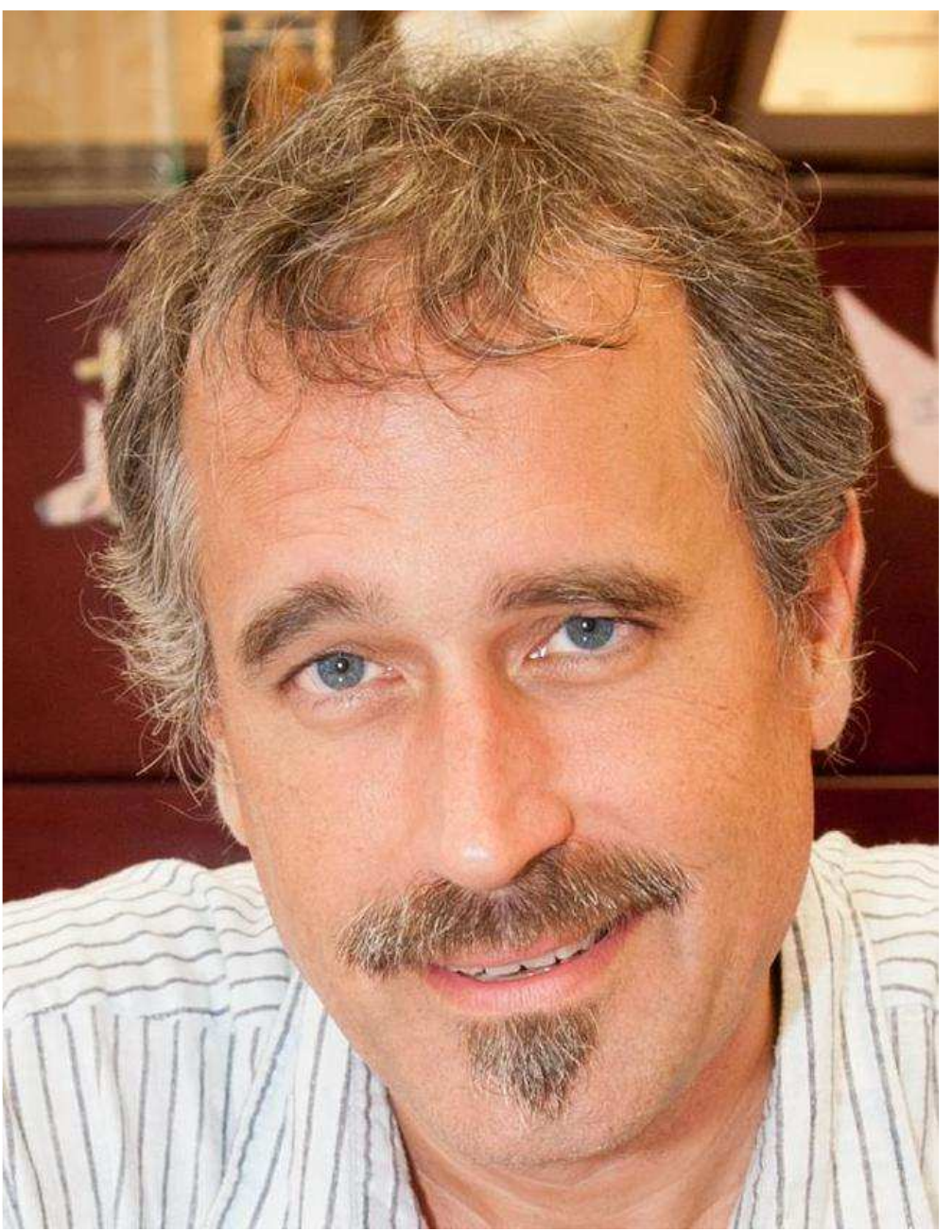}}]{Alan C. Bovik} holds the Cockrell Family Endowed Regents Chair in Engineering at The University of Texas at Austin, where he is Director of the Laboratory for Image and Video Engineering (LIVE) in the Department of Electrical and Computer Engineering and the Institute for Neuroscience. His research interests include image and video processing, digital television and digital cinema, computational vision, and visual perception. He has published over 750 technical articles in these areas and holds several U.S. patents. His publications have been cited more than 45,000 times in the literature, his current H-index is about 75, and he is listed as a Highly-Cited Researcher by Thompson Reuters. His several books include the companion volumes \emph{The Essential Guides to Image and Video Processing} (Academic Press, 2009).

Dr. Bovik received a Primetime Emmy Award for Outstanding Achievement in Engineering Development from the the Television Academy in October 2015, for his work on the development of video quality prediction models which have become standard tools in broadcast and post-production houses throughout the television industry. He has also received a number of major awards from the IEEE Signal Processing Society, including: the Society Award (2013); the Technical Achievement Award (2005); the Best Paper Award (2009); the Signal Processing Magazine Best Paper Award (2013); the Education Award (2007); the Meritorious Service Award (1998) and (co-author) the Young Author Best Paper Award (2013). He also was named recipient of the Honorary Member Award of the Society for Imaging Science and Technology for 2013, received the SPIE Technology Achievement Award for 2012, and was the IS\&T/SPIE Imaging Scientist of the Year for 2011. He is also a recipient of the Joe J. King Professional Engineering Achievement Award (2015) and the Hocott Award for Distinguished Engineering Research (2008), both from the Cockrell School of Engineering at The University of Texas at Austin, the Distinguished Alumni Award from the University of Illinois at Champaign-Urbana (2008). He is a Fellow of the IEEE, the Optical Society of America (OSA), and the Society of Photo-Optical and Instrumentation Engineers (SPIE), and a member of the Television Academy, the National Academy of Television Arts and Sciences (NATAS) and the Royal Society of Photography.

Professor Bovik also co-founded and was the longest-serving Editor-in-Chief of the \emph{IEEE Transactions on Image Processing} (1996-2002), and created and served as the first General Chair of the \emph{IEEE International Conference on Image Processing}, held in Austin, Texas, in November, 1994. His many other professional society activities include: Board of Governors, IEEE Signal Processing Society, 1996-1998; Editorial Board, \emph{The Proceedings of the IEEE}, 1998-2004; and Series Editor for Image, Video, and Multimedia Processing, Morgan and Claypool Publishing Company, 2003-present. His was also the General Chair of the 2014 Texas Wireless Symposium, held in Austin in November of 2014.

Dr. Bovik is a registered Professional Engineer in the State of Texas and is a frequent consultant to legal, industrial and academic institutions.

\end{IEEEbiography}

\end{document}